\documentclass{article}
\pdfoutput=1 

\usepackage[preprint, nonatbib]{NIPS}
\usepackage{xcolor}         
\usepackage[utf8]{inputenc} 
\usepackage[T1]{fontenc}    
\usepackage{hyperref}       
\usepackage{url}            
\usepackage{booktabs}       
\usepackage{amsfonts}       
\usepackage{nicefrac}       
\usepackage{microtype}      
\usepackage{graphicx}
\usepackage{caption}
\usepackage{amsmath}
\usepackage{subcaption}
\usepackage{pifont}
\usepackage[title]{appendix}
\title{Human-M3: A Multi-view Multi-modal Dataset for 3D Human 
Pose Estimation in Outdoor Scenes}

\author{Bohao Fan \thanks{Corresponding author.} \quad Siqi Wang \quad Wenxuan Guo \quad Wenzhao Zheng \quad Jianjiang Feng  \quad Jie Zhou \\
Beijing National Research Center for Information Science and Technology, China \\
Department of Automation, Tsinghua University, China \\
\texttt{\{fbh19,wangsq20,gwx22,zhengwz18\}@mails.tsinghua.edu.cn;} \\
\texttt{\{jfeng, jzhou\}@tsinghua.edu.cn}
}

\hypersetup{
colorlinks=true,
linkcolor=red,
citecolor=green
}

\begin{document}

\maketitle

\begin{abstract}
3D human pose estimation in outdoor environments has garnered increasing attention recently. However, prevalent 3D human pose datasets pertaining to outdoor scenes lack diversity, as they predominantly utilize only one type of modality (RGB image or pointcloud), and often feature only one individual within each scene. This limited scope of dataset infrastructure considerably hinders the variability of available data. In this article, we propose Human-M3, an outdoor multi-modal multi-view multi-person human pose database which includes not only multi-view RGB videos of outdoor scenes but also corresponding pointclouds. In order to obtain accurate human poses, we propose an algorithm based on multi-modal data input to generate ground truth annotation. This benefits from robust pointcloud detection and tracking, which solves the problem of inaccurate human localization and matching ambiguity that may exist in previous multi-view RGB videos in outdoor multi-person scenes, and generates reliable ground truth annotations. Evaluation of multiple different modalities algorithms has shown that this database is challenging and suitable for future research. Furthermore, we propose a 3D human pose estimation algorithm based on multi-modal data input, which demonstrates the advantages of multi-modal data input for 3D human pose estimation. Code and data will be released on \url{https://github.com/soullessrobot/Human-M3-Dataset}.

\end{abstract}

\section{Introduction}

In recent years, more and more attention has been paid to 3D human pose estimation (HPE) algorithms \cite{lin2021multi, zhang2021direct, moon2018v2v, reddy2021tessetrack, tu2020voxelpose, li2022lidarcap, zhao2022lidar}. These algorithms use multiview RGB images or LiDAR pointclouds as inputs to estimate the 3D pose and shape of the human body. 3D HPE is the foundation for human action recognition, social behavior analysis, scene perception, and analysis of many downstream applications, including augmented/virtual reality, simulation, and autonomous driving.

\begin{figure*}
    \includegraphics[width=1.0\textwidth]{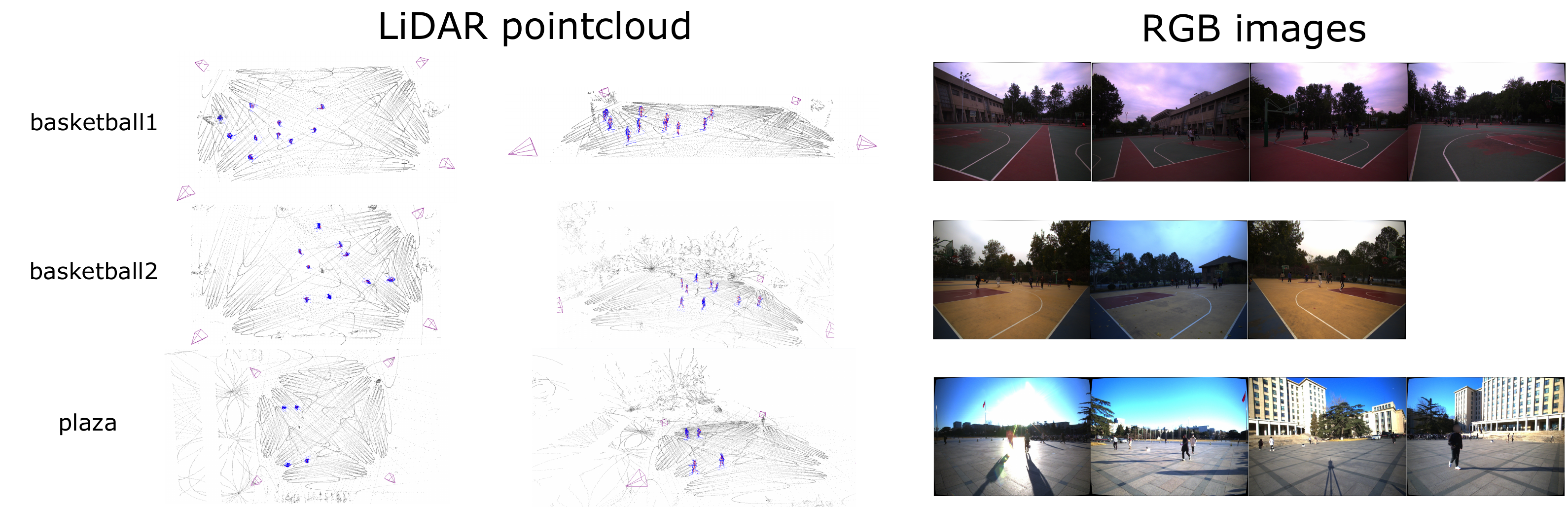}
    \vspace{-3mm}
    \caption{Examples of pointcloud and RGB images collected in three different scenes. Each row shows a distinct scene, from top to bottom are "basketball1", "basketball2" and "plaza" respectively. First column shows bird eye view of pointcloud and annotations. Gray points indicate background points and blue points are foreground points from human surface segmented by groundtruth. Groundtruth 3D human poses are marked with red lines and purple pyramid indicates the location of capture units. Second column shows another view of them. the last four columns show four RGB images captured simultaneously. It is worth mentioning that "basketball2" scene only possesses 3 capture units.}
\label{fig:scene}
\vspace{-6mm}
 \end{figure*}
\begin{figure*}
  \centering
  \includegraphics[width=0.8\textwidth]{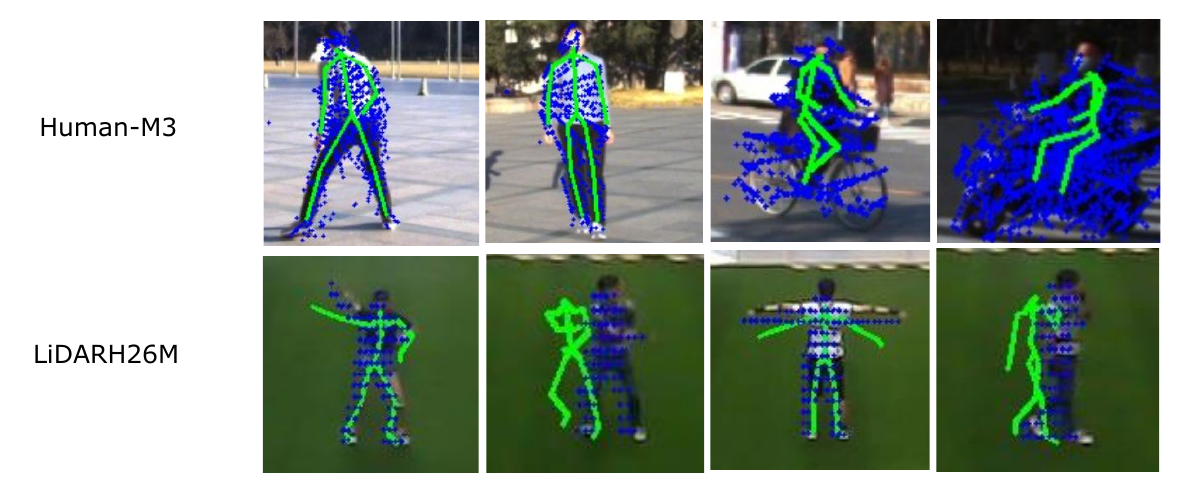}
  \vspace{-3mm}
  \caption{Comparison of human pose annotations from Human-M3 and LiDARH26M \cite{li2022lidarcap} dataset. First row are examples from the proposed Human-M3 dataset. Second row are examples from LidarH26M dataset, where ground truth poses are not well aligned with images and pointclouds. }
  \label{fig:comparison_data}
  \vspace{-6mm}
\end{figure*}

  
Over the past few decades, several datasets and benchmarks have been proposed, greatly advancing research in 3D HPE. These datasets can be classified into two categories based on the collection scenes. The first category typically uses optical tracking systems \cite{ionescu2013human3, sigal2010humaneva, mahmood2019amass}, camera arrays \cite{joo2015panoptic, weinzaepfel2020dope, chen2021anatomy}, and RGB-D sensors \cite{hassan2019resolving, zhang2022egobody} to capture partial body poses under constrained conditions. However, optical systems are sensitive to natural light and lack depth information, making them unstable in outdoor scenes, and space-time synchronization between camera arrays brings physical limitations to the acquisition system, which also cannot be applied to outdoor scenes. RGB-D sensors are constrained by a limited range and are not suited for outdoor applications. The second category attempts to capture unobstructed 3D poses in outdoor environments using IMU sensors mounted on the human body. However, IMUs are prone to drift in long-term acquisition processes, resulting in misalignment with the human body. As a result, some methods use additional sensors such as RGB cameras \cite{trumble2017total, xu2017flycap, guzov2021human}, RGB-D cameras \cite{zheng2018hybridfusion}, or LiDAR\cite{li2022lidarcap, dai2023sloper4d} to provide constraints and achieve significant improvements. In particular, Dai et al. \cite{dai2023sloper4d} used the interaction between the pedestrian and the environment as an additional constraint. However, due to the limitations of motion capture devices, these scenarios are restricted to single person and lack interaction between different individuals, severely limiting the diversity of captured data. As illustrated in Fig. \ref{fig:comparison_data}, registration between local 3D human poses collected from motion capture equipment and pointcloud is problematic in LiDARH26M \cite{li2022lidarcap} dataset. Furthermore, the majority of these scenarios only pertain to solitary individuals.

We consider the primary obstacle in acquiring 3D human pose data for outdoor scenes to be:

1) Obtaining accurate ground truth poses directly is challenging in large outdoor scenes, particularly for humans who cannot be equipped with motion capture devices.
2) Pedestrians often appear small on images due to their distance from sensors, posing a challenge for pose estimation.
3) Pedestrians are frequently occluded by intervening individuals or objects, rendering them difficult to discern.

\begin{table}
\small
\caption{Comparison with existing 3D human pose databases. "Num" denotes number of valid 3D human poses, "I" denotes image and "P" denotes pointcloud.}
\begin{center}
\begin{tabular}{ c | c c c c c c } 

\hline
\textbf{Dataset} & \textbf{Num} & \textbf{Modality} & \textbf{Multi-view} & \textbf{Multi-Person} & \textbf{Multiple Scenes} & \textbf{Scene} \\
\hline
Human 3.6M \cite{ionescu2013human3} & 3.6M & I & \checkmark & \ding{53} & \ding{53} & Indoor \\
3DPW \cite{von2018recovering} & 50.1K & I & \ding{53} & \checkmark & \checkmark & Outdoor \\
PedX \cite{kim2019pedx} & 10.1K & P & \ding{53} & \checkmark & \checkmark & Outdoor \\
LiDARH26M \cite{li2022lidarcap} & 180.4K & P & \ding{53} & \ding{53} & \ding{53} & Indoor \\
SLOPER4D \cite{dai2023sloper4d} & 100K & I+P & \ding{53} & \ding{53} & \checkmark & Outdoor \\
Human-M3 & 89.6K & I+P & \checkmark & \checkmark & \checkmark & Outdoor \\
\hline
\end{tabular}
\end{center}
\label{table:tabledatacompare}
\vspace{-7mm}
\end{table}

To tackle the aforementioned challenges, we propose a novel approach that leverages multiple sets of cross-view capturing units for outdoor scene acquisition. Following \cite{zhang2022flexible}, each unit comprises a LiDAR and a camera, enabling pedestrian detection and pose estimation without requiring motion capture devices. Leveraging the acquired multi-modal dense information, our proposed pipeline offers a robust method for accurate 3D HPE, spanning from pedestrian localization to 3D pose estimation. In order to ensure the validity of the ground truth poses, we meticulously select and reconstruct erroneous poses, thereby obtaining precise and impactful human pose data for outdoor scenes.

We present a novel dataset, tailored to 3D HPE tasks in outdoor environments. As illustrated in Fig. \ref{fig:scene}, this dataset comprises a combination of multi-view images and LiDAR data, which provides a unique and unparalleled resource for 3D HPE. As shown in Table \ref{table:tabledatacompare}, this dataset has the capability to capture multiple individuals within outdoor environments simultaneously, thereby expanding the range of scenes exemplifying interactions between multiple people. Human-M3 offers a comprehensive representation of human pose and the surrounding environment, making it more suitable for advancing the state-of-the-art in 3D HPE research. Furthermore, in order to comparably examine the potential strengths and limitations of distinct modalities for 3D HPE in our Human-M3 database, we conducted a series of empirical analyses, ultimately leading us to propose a multi-modal approach for human body pose estimation in order to showcase the advantages of incorporating diverse data modalities.

The principal contributions of this article can be summarized as follows:

1) We present the first comprehensive 3D human pose dataset that integrates multi-camera and LiDAR in outdoor scenes.
2) We develop an efficient method for collecting 3D human body pose data of multiple individuals in outdoor settings, without the need for wearable devices.
3) We introduce a 3D human pose estimation algorithm that effectively integrates multi-modal data inputs, emphasizing the importance and effectiveness of utilizing such multi-modal data.

\section{Related work}
\subsection{3D human pose datasets}
State-of-the-art 3D human pose databases can can be classified into distinct groups based on their environmental classification, including either indoor or outdoor datasets.

\textbf{Indoor datasets.}
There is a diverse array of datasets available for researching 3D HPE, utilizing a range of RGB sensors and configurations. Among them, Human3.6M \cite{ionescu2013human3} dataset contains vast indoor motion-captured recordings of human movements, PROX\cite{hassan2019resolving} dataset uses RGB-D sensors to capture single-person video data in various indoor environments. Panoptic \cite{joo2015panoptic} dataset comprehensively captures indoor scenes with many spatially and temporally calibrated RGB-D sensors, and ITOP \cite{haque2016towards} dataset is a depth-based database that combines different depth camera setups with motion capture devices to obtain human poses. These existing datasets confine to indoor environments, lack variations in pedestrian interactions, and do not account for outdoor scenarios.

\textbf{Outdoor datasets.}
The 3DPW \cite{von2018recovering} is the pioneering database that offers 3D annotations for outdoor environments using a handheld RGB camera, with annotations obtained via a registration process involving 2D poses and 3D poses obtained through motion capture. Similarly, the HPS \cite{guzov2021human} dataset adopted a similar approach to reconstruct a broader range of outdoor scenarios. PedX \cite{kim2019pedx} is the initial outdoor pose dataset that does not rely on motion capture equipment. It captured a comprehensive range of street scenes, and also utilized image-based ground truth labels to estimate body meshes. However, the reliance on LiDAR sensors may cause issues with pose estimation accuracy due to sparse data when individuals are too far from the LiDAR. Additionally, mesh estimation quality from images may not be ideal. Li et al. \cite{li2022lidarcap} used a single LiDAR to collect precise pointcloud, while ground truth poses were acquired through carefully calibrated motion capture equipment within a range of 15-30 meters from the sampled individual. It only included single individuals and their interactions with others were not effectively captured. Dai et al. \cite{dai2023sloper4d} presented a novel approach for data collection, in which individuals wear LIDAR devices and detection equipment, significantly enhancing the size and diversity of scenes captured. However, they can only effectively capture a limited number of individuals wearing motion capture equipment in the same scene. Ultimately, and fail to capture multi-person human poses and interactions effectively without reliance on motion capture equipment, resulting in a lack of complexity in multi-person scenarios. 

The Human-M3 database can serve as a valuable benchmark for assessing the performance of outdoor multi-person 3D HPE algorithms. Additionally, the multisensory characteristics of the database could facilitate the evaluation of algorithms that incorporate multiple modalities.
\subsection{3D HPE algorithms}
\textbf{Multi-view 3D HPE algorithms} leverage synchronized inputs from multiple RGB cameras to determine the 3D human pose. They are mainly facilitated by enforcing geometric constraints between input images. These methods first process each input image to obtain features or 2D poses of the human body then match and fuse them in 3D space. In addition to estimate local human pose and orientation, these algorithms also generate the precise global position of each individual in the world coordinate system. Dong et al. \cite{dong2019fast} proposed a method based on singular value decomposition to solve the matching problem between multi-view human poses. Tu et al. \cite{tu2020voxelpose} leveraged projection relationships to fuse keypoint heatmaps from multiple views into 3D space, before utilizing a 3D convolutional neural network to estimate the 3D pose. Lin et al. \cite{lin2021multi} employed 2D detections from each view to align and aid the estimation of 2D depth for each person, and then fused these estimations across multiple views to produce a reliable 3D pose. Zhang et al. \cite{zhang2021direct} utilized the Vision Transformer architecture to establish feature connections between multiple views, leveraging transformers to achieve effective human pose regression.

\textbf{Pointcloud-based 3D HPE algorithms} utilize one or more frames of single-person pointclouds as input to estimate the body pose. These algorithms typically operate under the assumption that the absolute position of the humans have already been located, and aim to regress local human poses. Moon et al. \cite{moon2018v2v} voxelized the 3D pointcloud and utilized 3D convolution to regress heatmaps, which were then used to extract 3D human pose heatmaps. Li et al. \cite{li2022lidarcap} used person pointcloud videos as input and employed the structure of PointNet++ cascaded with GRU to regress the 3D human pose, surpassing RGB video input in LiDAR scenes.

All aforementioned algorithms are based on single-modalities, meaning they utilize only one type of data input. Currently, there are no 3D HPE algorithms that combine pointclouds with RGB images to the best of our knowledge.

\section{Approach}
\subsection{Data acquisition}
We used similar devices and scene setups as described in \cite{zhang2022flexible}. Each capture unit combines a LiDAR and a camera, which are synchronized in both space and time. The Human-M3 dataset consists of 12,200 frames captured from four different types of scenes, including intersection, plaza, and two types of basketball courts. Fig. \ref{fig:scene} shows examples from three of them, with each frame composed of 3-4 capture units. While the basketball court scenes typically feature ten players, the intersection and plaza scenes also include random pedestrians, cyclists, and bikers. The Human-M3 database provides a comprehensive set of 89642 valid human pose records for assessment purposes.

\subsection{Data annotation}
\label{sec:data_anno}
\begin{figure*}
  \centering
  \includegraphics[width=0.9\textwidth]{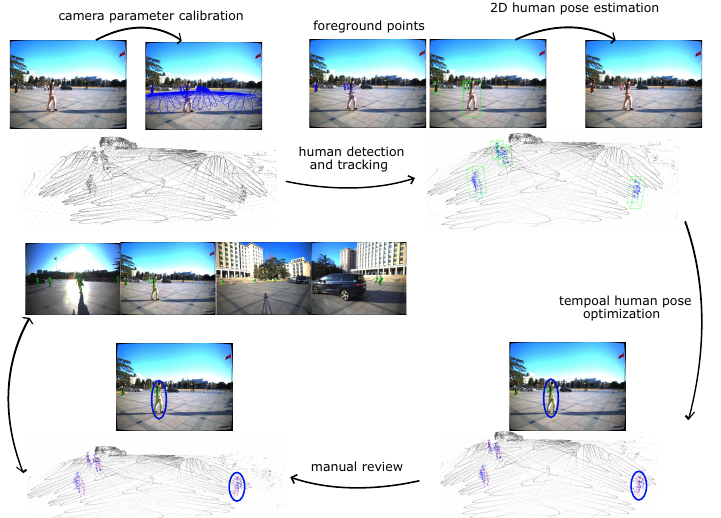}
  \caption{The overall structure for data annotation process.}
  \label{fig:pipeline}
  \vspace{-4mm}
\end{figure*}

The overall structure of the data annotation process is illustrated in Fig. \ref{fig:pipeline},  after receiving the raw multi-modal data, the processing steps can be divided into four steps: human detection and tracking, temporal human pose optimization and manual review.

\textbf{Human detection and tracking.} We first calibrate the extrinsic camera parameters by manually annotating corresponding point pairs between RGB images and pointcloud. More detailed descriptions of this step are given in supplementary materials. We then utilized PointPillars \cite{lang2019pointpillars} for pedestrian detection from input pointclouds and AB3DMOT \cite{weng2020ab3dmot} for tracking. In this step, we manually annotated 12.5\% of the data as the training set to train the model, and utilized them to estimate the remaining data. During this process, the final bounding box was filtered based on the number of LiDAR points inside, eliminating false detection. Upon completing manual counting process, a total of 961 detections were ultimately lost, representing 0.86\% of the overall number of occurred samples. Then we projected the 3D bounding box onto images to obtain 2D bounding box for each person in every camera view. Subsequently, we utilized a top-down pose detector ViTPose \cite{xu2022vitpose} to obtain precise human poses. We found that this methodology yields pixel-level accurate 2D pose estimation results without necessitating additional training or multi-view registration. 

\textbf{Temporal human pose optimization.}
We propose an optimization-based method for multi-person pose estimation in complex scenarios. In the Panoptic \cite{joo2015panoptic} dataset, a spatially dense camera array was utilized to perform a bottom-up 2D pose estimation using Openpose \cite{cao2017realtime}. The resulting 2D poses were then reconstructed into 3D poses using multi-view epipolar geometry constraints. This methodology has been proven to be robust when the camera array possesses high density. However, obtaining a dense camera array for outdoor scenes is challenging. Inspired by HuMoR \cite{rempe2021humor} and Smplify-x \cite{pavlakos2019expressive}, we propose a similar methodology that uses sparse multi-view 2D pose estimation and pointclouds to reconstruct ground truth in 3D poses. To achieve better performance, we propose a temporal optimization algorithm that optimizes the 3D pose of each person distinctively, which makes use of constraints derived from pointclouds and temporal optimization.

For a pedestrian locating in $R = [R_0, R_1, ..., R_{t-1}]$ with $t$ frames, we try to optimize SMPL \cite{loper2015smpl} shape parameter $\beta$, SMPL pose parameters $\theta=[\theta_0, \theta_1, ..., \theta_{t-1}]$ and root translation $r=[r_0, r_1, ..., r_{t-1}]$ with the estimated 2D human pose $P=[P_{0}, P_{1}, ..., P_{t-1}]$ and corresponding local pointclouds $Q=[Q_{0}, Q_{1}, ..., Q_{t-1}]$. ($P_t = [P_{0t}, P_{1t}, ..., P_{C-1t}]$ and $C$ is number of cameras.) This problem can be formulated as minimizing the loss function constraining observations and priors:
\begin{equation}
    [r_o,\beta_o,\theta_o] = \mathop{\arg\min}\limits_{r, \beta, \theta}(L(r,\beta, \theta, P, Q))
\end{equation}
\begin{equation}
    L(r,\beta, \theta, P, Q) = L_{D}(r, \beta, \theta, P, Q) + L_{sp}(\beta) + L_{pp}(\theta) + L_{mp}(\theta)
\end{equation}
\begin{equation}
L_{D}(r, \beta, \theta, P, Q) = L_{I}(r, \beta, \theta, P) + L_{P}(r, \beta, \theta, Q)
\end{equation}
$L_{I}$ constraints human joints to be consistent with the 2D HPE results from each view. $L_{P}$ is the chamfer loss that constrains mesh points and corresponding pointclouds as close as possible. $L_{sp}$ and $L_{pp}$ are human shape prior and human pose prior inherited from VPoser \cite{pavlakos2019expressive}. $L_{mp}$ restricts the dynamic change of human pose from being too drastic. The initial value of $r$ is set to $R$ and others are set to 0. The optimization can be achieved by L-BFGS algorithm and more detailed description of this step is given in supplementary materials.

The principal aim of the algorithm is to uphold prior knowledge and pose smoothing, while maximizing adherence to each individual's 2D HPE outcomes and pointcloud observations. After optimizing with suitable hyperparameters, we obtain satisfactory 3D pose sequences $[r_o,\beta_o,\theta_o]$. Further, we feed them into SMPL model to obtain 3D keypoints $J=[J_0,...,J_{t-1}]$.

\textbf{Manual review.}
Due to occlusion and incorrect 2D HPE, suboptimal 3D human poses may be produced during the aforementioned processes. We project 3D keypoints onto each camera view to synthesize video footage, and proceeded to meticulously scrutinize individual frames for incorrect human poses. Any frames depicting suboptimal human poses were then either updated through interpolation with the correct human pose seen in the preceding and following frames or removed when irretrievable. During this process, we removed 10442 human poses (10.43\% of total detected humans) to ensure the validity of annotations.

\subsection{A simple baseline for multi-modal 3D HPE}
\begin{figure*}
  \centering
  \includegraphics[width=1.0\textwidth]{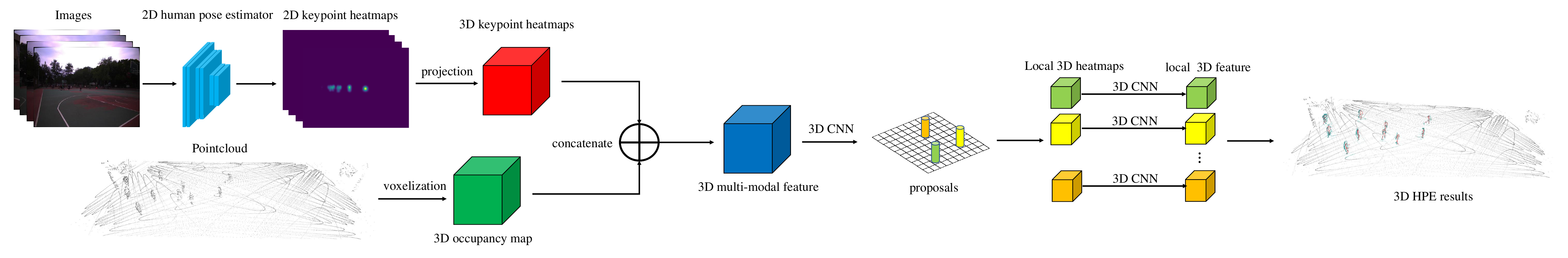}
  \vspace{-3mm}
  \caption{The overall structure of the proposed 3D HPE algorithm. }
  \label{fig:structure}
  \vspace{-5mm}
\end{figure*}

We propose a simple voxel-based approach for pedestrian detection and human pose estimation that integrates both RGB image and pointcloud inputs, namely MultiModal-VoxelPose (MMVP). The overall structure of MMVP is illustrated in Fig. \ref{fig:structure}. We utilized a 2D human pose estimator OpenPIFPAF \cite{kreiss2021openpifpaf} to generate 2D keypoint heatmaps from RGB images, which were transformed into 3D heatmaps using 2D-3D projection. Meanwhile, the pointcloud input was voxelized and sampled into a 3D occupancy volume. These two sets of features were subsequently fused and fed into a 3D CNN for the extraction of 3D features and regression of 3D heatmaps. It is worth mentioning that the fusion part can use any kind of fusion module, and here we use the simplest of cascades. The 3D CNN part and corresponding losses of the method are similar to that in VoxelPose \cite{tu2020voxelpose}.

\section{Experiments}
\subsection{Experimental settings and metrics}
\textbf{Data setting.} For evaluation, the four data segments in the database were partitioned into two parts. The first 90\% of the data in each segment was designated as the training set, while the remaining 10\% was designated as the test set.

\textbf{Algorithms.} We evaluated the MMVP in the Human-M3 dataset, to more intuitively illustrate the effectiveness of multimodal data and algorithms, we evaluated the performance of several state-of-the-art multiple-view 3D HPE algorithms including Voxelpose \cite{tu2020voxelpose}, PlaneSweepPose \cite{lin2021multi}, and MVP \cite{zhang2021direct}, as well as the pointcloud-based 3D HPE algorithm V2V-PoseNet\cite{moon2018v2v}. We report Mean Per Joint Position Error (MPJPE), Recall with a threshold of 500 mm, and Average Precision (AP) with thresholds ranging from 75 mm to 150 mm with 25mm step. 

\textbf{Implementation details.} We train MMVP on 4 TITAN RTX GPUs and the overall batch size is set to 4. We use Adam \cite{kingma2014adam} optimizer for training and the learning rate was set to 0.0001, where we train the model for 20 epochs to converge. Models of compared algorithms were also trained until converge. To achieve fairness, we employed a consistent approach for the input of VoxelPose and PlaneSweepPose. Specifically, the estimated heatmaps from OpenPIFPAF \cite{kreiss2021openpifpaf} serve as inputs for both methods, which are aligned with the input adopted by MMVP. Furthermore, as V2V-PoseNet is a 3D HPE algorithm designed for single-person and does not integrate a pedestrian localization algorithm, we adapted the cuboid proposal network in VoxelPose. Specifically, we modified the input of V2V-PoseNet into a voxelized pointcloud, which served as the localization component.

\begin{figure*}
  \centering
  \begin{subfigure}{0.24\linewidth}
  \includegraphics[width=1.0\textwidth]{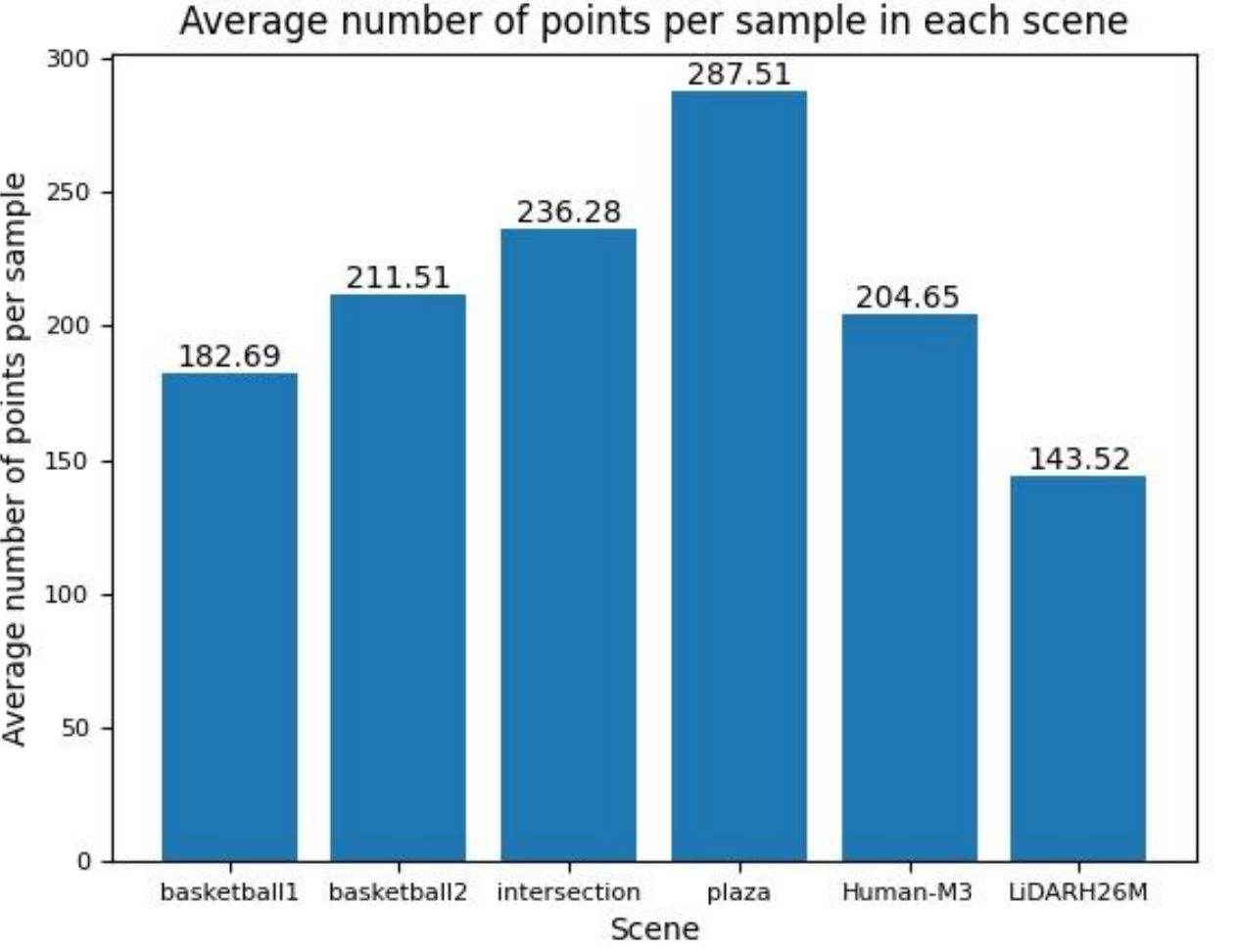}
  \caption{}
    \label{fig:pointnumber}
  \end{subfigure}
  \begin{subfigure}{0.24\linewidth}
  \includegraphics[width=1.0\textwidth]{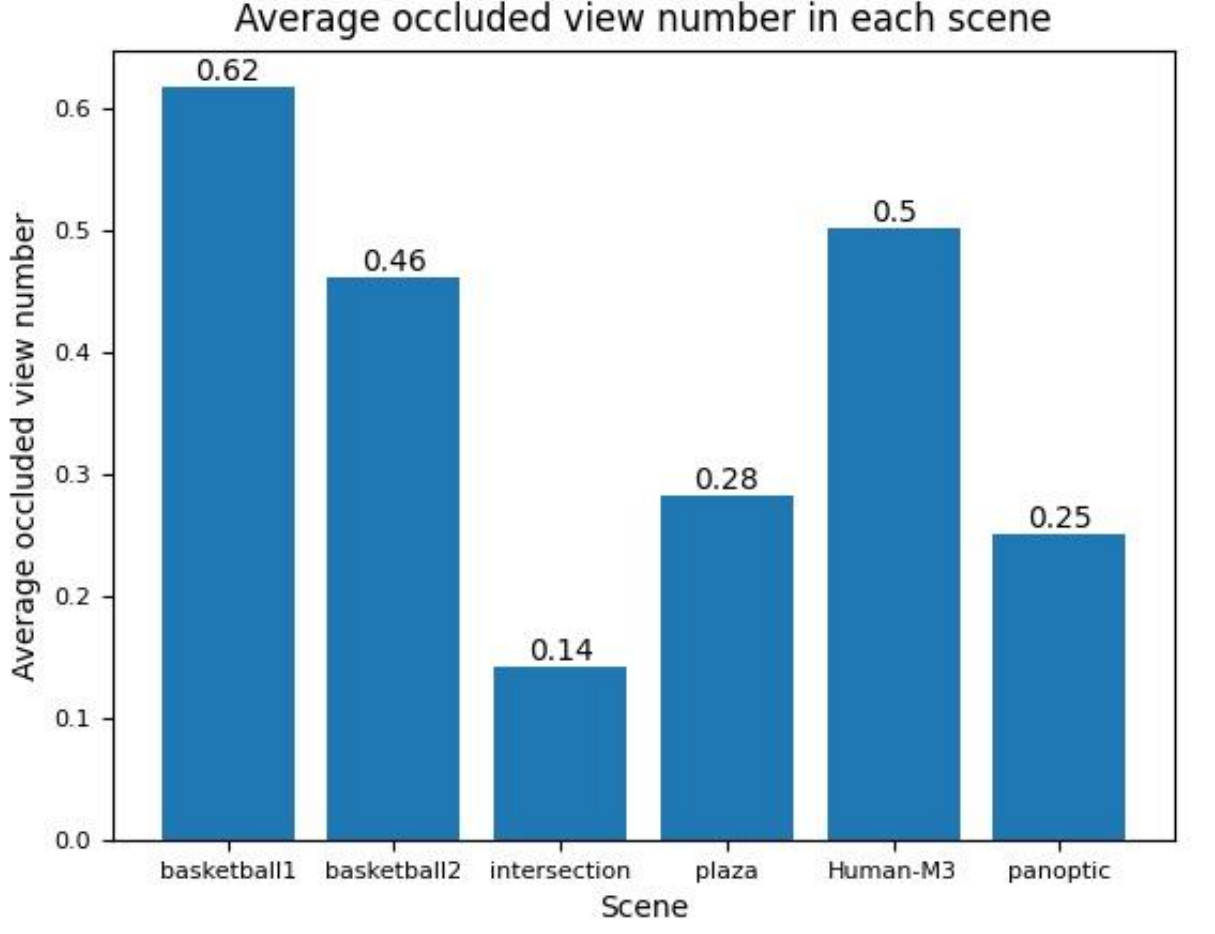}
  \caption{}
    \label{fig:occnumber} 
  \end{subfigure}
 \begin{subfigure}{0.24\linewidth}
  \includegraphics[width=1.0\textwidth]{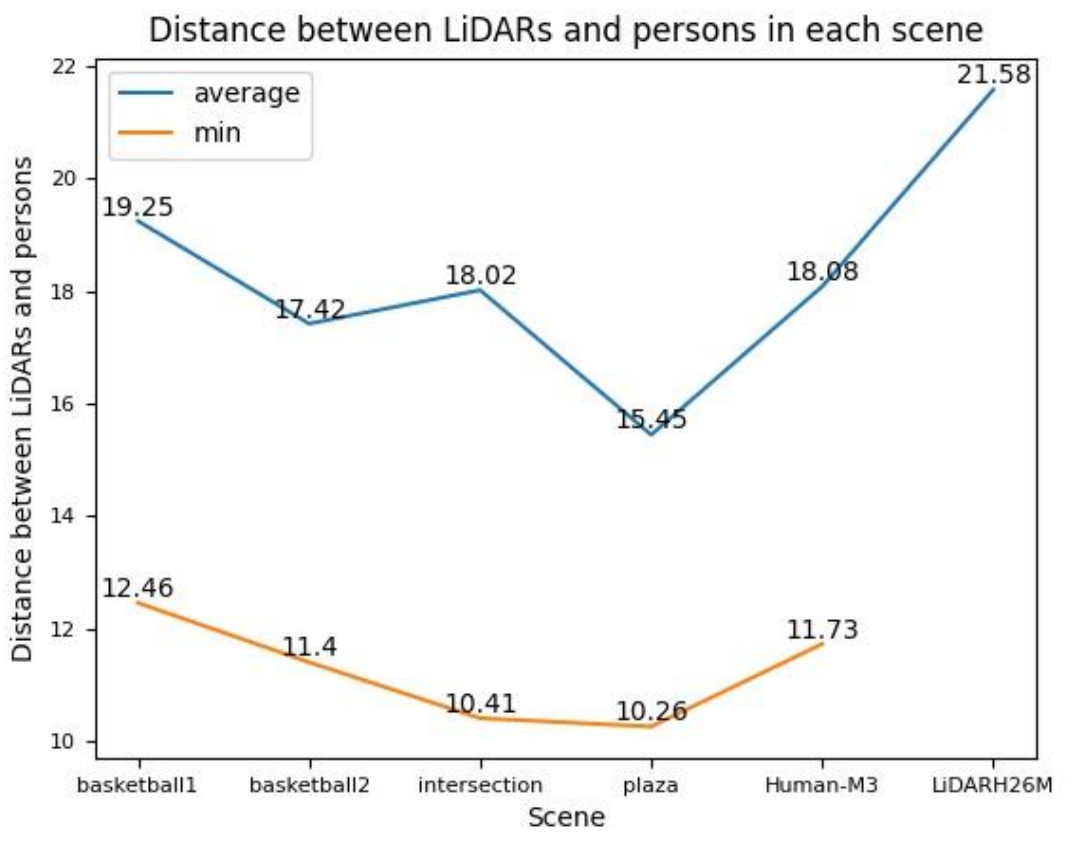}
  \caption{}
    \label{fig:distance}
  \end{subfigure}
  \begin{subfigure}{0.24\linewidth}
  \includegraphics[width=1.0\textwidth]{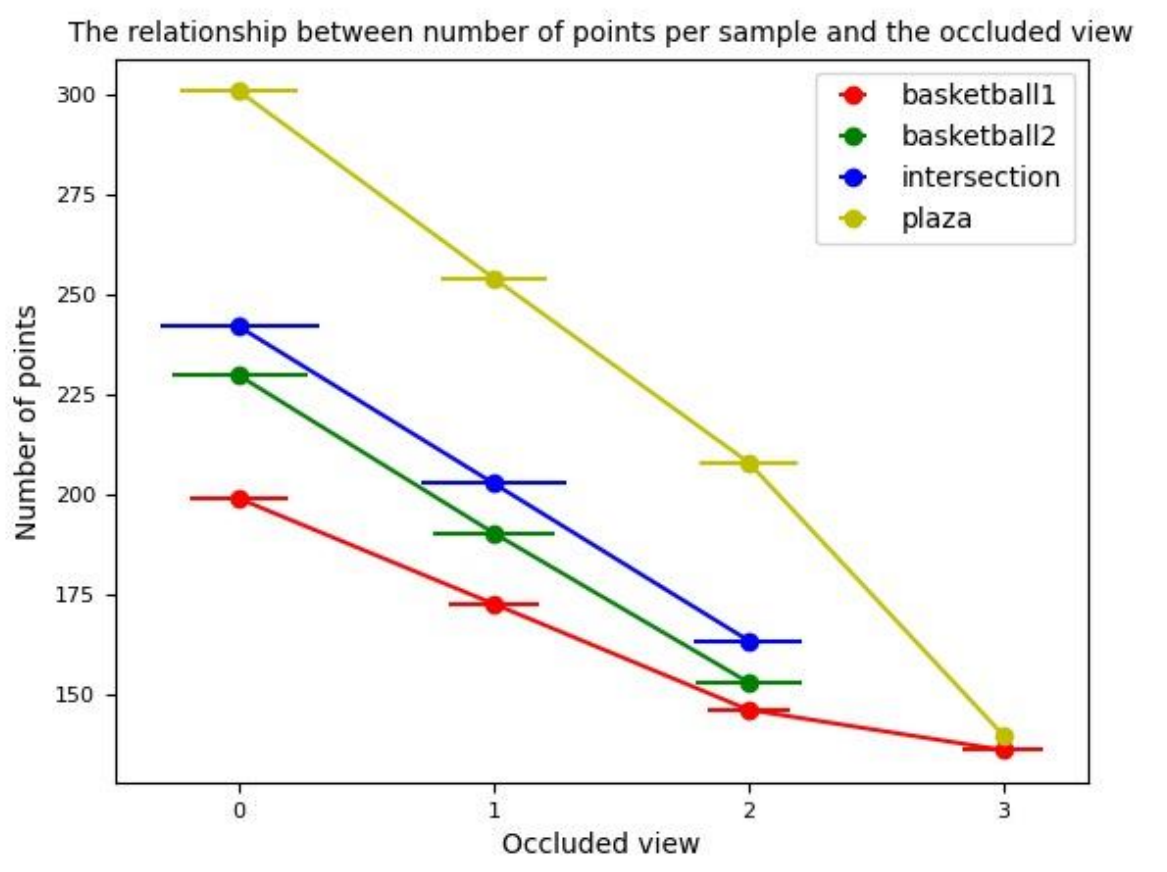}
  \caption{}
    \label{fig:distance_number} 
  \end{subfigure}
  \vspace{-3mm}
  \caption{Analysis of several data characteristics of Human-M3 dataset. (a) is the statistical chart of average point number in each scene. (b) the average occluded view number in each scene. (c) the distance between LiDARs and persons in each scene. Given the multiple LiDAR Settings in the scene, "average" means the average distance between persons and all LiDARs, and "min" means the distance between persons and their nearest LiDAR. (d) the relationship between point number and occluded view number. The length of horizontal lines represents the value of the statistical variance.}
  \label{fig:data_characteristic}
  \vspace{-4mm}
\end{figure*}

\subsection{Data characteristic}
As illustrated in Table \ref{table:tabledatacompare}, the Human-M3 database offers a slightly lower count of 3D human poses as compared to the LiDARH26M \cite{li2022lidarcap} and SLOPER4D  \cite{dai2023sloper4d} databases. Nevertheless, the combination of multi-view and multi-person scene setup still offers sufficient richness. We compare the Human-M3 dataset with Panoptic \cite{joo2015panoptic} and LiDARH26M \cite{li2022lidarcap} datasets for analogous scene settings. The Panoptic dataset aims to provide precise human poses in multi-view and multi-person scenarios using keypoint coordinates, while LiDARH26M provides human pointcloud data. We focus on some data characteristics and compare them with the aforementioned two single-modal database.

\textbf{Number of points per sample.} As illustrated in Fig. \ref{fig:pointnumber},  the Human-M3 dataset contains an average of 200 points per person, slightly exceeding that of the LiDARH26M dataset. Due to the surrounding placement of the sensors, the Human-M3 dataset provides more comprehensive semantic data for 3D human pose estimation than the LiDARH26M dataset.

\textbf{Number of views occluded.} Fig. \ref{fig:occnumber} compares the 2D occlusion between humans in Human-M3 and Panoptic datasets. We consider a pose to be occluded if its bounding box is occluded by other bounding boxes from individuals closer to the camera by more than 50\% of their bounding boxes. The Human-M3 dataset exhibits a slightly higher proportion of occluded poses, suggesting that monocular or multi-view 3D human pose estimation tasks may face more challenging scenarios. Additionally,  the LiDAR struggles to capture complete information accurately when pedestrians are severely obstructed in various viewpoints as shown in Fig. \ref{fig:distance_number}. Occlusions reduce the number of points available and make pointcloud analysis more challenging.

\textbf{Distance to LiDARs.} Persons in the Human-M3 dataset are closer to LiDARs than those in the LiDARH26M dataset as illustrated in Fig. \ref{fig:distance}, but the nearest LiDAR in the scene is still more than 10 meters away from individuals on average, thereby preserving the outdoor scene's characteristics. 

The integration of multi-view images and LiDAR pointclouds represents a paradigm shift in 3D information fusion by leveraging the unique modalities of each. The sparsity of cameras in the scene results in a diverse environment for pedestrians both in close proximity and far away, offering potential benefits for the training of single-view pose estimation algorithms. As far as we know, this is the first attempt to present multi-view image data captured in outdoor scenarios featuring multiple individuals engaged in interactive activities. Additionally, the broad overlap range of multiple LiDARs can be utilized to guide 3D HPE algorithms that utilize sparse LiDAR data.

\begin{table}
\small
\caption{Comparison with state-of-the-art 3D HPE algorithms. Metrics include MPJPE (m), recall (500mm) and average precision (AP). Best values are shown with blod. (PCD: pointcloud input.)}
\begin{center}
\begin{tabular}{ c | c c c c c c c} 
\hline
\textbf{Algorithm} & \textbf{Input Modality} & \textbf{MPJPE} & \textbf{Recall} & $AP_{75}$ & $AP_{100}$ & $AP_{125}$ & $AP_{150}$ \\
\hline
VoxelPose & RGB & 0.108 & 90.82 & 22.97 & 44.85 & 59.65 & 68.49 \\
PlaneSweepPose & RGB & 0.099 & 39.78 & 4.66 & 8.43 & 10.85 & 12.09 \\
MVP & RGB & 0.14 & 89.26 & 7.44 & 22.27 & 37.65 & 50.12  \\
V2V-PoseNet & PCD & 0.107 & 98.23 & 11.69 & 34.88 & 57.00 & 72.50 \\
MMVP(ours) & RGB + PCD & \textbf{0.079} & \textbf{98.31} &\textbf{41.92} & \textbf{69.85} & \textbf{81.50} & \textbf{87.65}  \\
\hline
\end{tabular}
\end{center}
\label{table:tablealgorithm}
\vspace{-7mm}
\end{table}

\begin{figure*}
  \centering
  \includegraphics[width=0.9\textwidth]{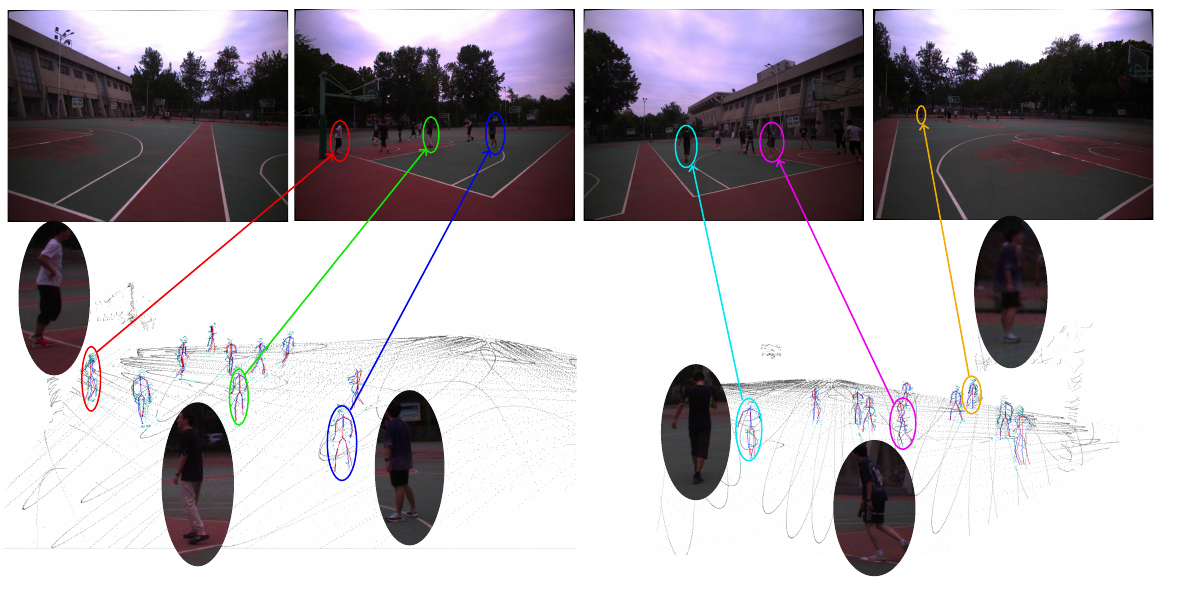}
  \vspace{-3mm}
  \caption{An exmple of MMVP results on the proposed Human-M3 dataset. In the second row, side views of the two scenes are shown separately. Inside them, red lines indicate predicted 3D human poses and blue lines indicate ground truth poses. Foreground points are marked as cyan points, and background points are gray. Corresponding RGB images are shown in the first row. We use ellipses and arrows to mark the correspondence for better understanding.}
  \label{fig:demo_show}
  \vspace{-5mm}
\end{figure*}

\subsection{Evaluation and discussion}
\textbf{Pointcloud vs multi-view RGB images for 3D HPE.} As illustrated in Table \ref{table:tablealgorithm}. We found that models based on pointclouds as input perform better than RGB images at localization and human pose estimation. Specifically, V2V-PoseNet outperforms VoxelPose with 1.4cm in terms of MPJPE and 7.32\% in terms of recall. Specifically, the 3D components of the two networks possess the same structure. Image-based algorithms tend to miss detection more frequently when compared with pointcloud-based algorithms.
This could be attributed to the fact that in a vast outdoor environment, individuals are typically located at a considerable distance from the camera, and occlusions occur more frequently. Consequently, 2D images may lack sufficient local details, making it challenging to accurately gauge the position of persons. On the contrary, information acquired from point clouds is less affected by distance attenuation and proved to be more reliable. A similar phenomenon has also occurred in the results of MVP and PlaneSweepPose. The results of these algorithms fall well short of their performance on the Panoptic database. This also reflects the characteristics and challenges of the Human-M3 dataset.

\textbf{Fusion method of multi-view 3D HPE algorithm.} PlaneSweepPose depends on detailed 2D human pose matching to estimate pedestrian depth from multiple viewpoints. In large scenes with pedestrians at a far distance, depth estimation becomes cumbersome. As reflected by the 39.41\% recall value, this kind of localization pipeline lacks universality and leads to a large number of omissions. Similarly, the MVP algorithm, whose method to locate people is to obtain the consistency of multiple perspectives through transformer. This requires precise 2D features, also hard to achieve optimal results in wide scenes. In contrast, the simplest and most intuitive voxel-based fusion method achieves better results. In addition, this fusion template is also more friendly to multi-modal fusion, according to the 3D spatial correspondence, pointclouds can be easily fused together as input.

\textbf{The role of multimodal inputs.}
When multi-modal data is used as input, simple concatenation of 3D heatmaps is enough to significantly improve algorithmic performance. In particular, the MPJPE remarkably decreased by 2.6cm (24\%) and AP in each threshold rised at least by 0.25. The role of multimodal data input is obvious. We claim that pointcloud and RGB images provide complementary information: pointcloud can provide the exact position of the pedestrian as well as scale-invariant features which are independent of human location, but cannot provide enough texture information; RGB features can provide more texture information, but are easily affected by camera distance and occlusion. Fusion of the them makes it easier for the model to extract reliable features from data, highlighting the value of multi-modality for 3D HPE. Two examples are illustrated in Fig. \ref{fig:demo_show}, human poses predicted by the MMVP algorithm are close to ground truth even though the persons are quite far from the sensors. More detailed visualization results are illustrated in supplementary materials.

\begin{table}
\small
\caption{MMVP results in different scenes. Metrics include MPJPE (m), recall (500mm) and average precision (AP).}
\begin{center}
\begin{tabular}{ c | c c c c c c c} 
\hline
\textbf{Scene} & \textbf{Number of poses} & \textbf{MPJPE} & \textbf{Recall} & $AP_{75}$ & $AP_{100}$ & $AP_{125}$ & $AP_{150}$ \\
\hline
basketball1 & 3821 & 0.0763 & 98.61 & 42.47 & 70.57 & 83.15 & 90.92 \\
basketball2 & 4751 & 0.0755 & 98.27 & 44.72 & 72.94 & 85.01 & 89.97 \\
intersection & 261 & 0.2262 & 77.39 & 0.0 & 0.0 & 0.15 & 2.54 \\
plaza & 480 & 0.0795 & 100.0 & 35.18 & 68.53 & 86.14 & 93.76 \\
\hline
overall & 9313 & 0.079 & 98.31 & 41.92 & 69.85 & 81.50 & 87.65 \\
\hline
\end{tabular}
\end{center}
\label{table:tablealgorithm_scene}
\vspace{-5mm}
\end{table}

\textbf{Evaluation between different scenes.}
As shown in Table \ref{table:tablealgorithm_scene}, we evaluated the performance of MMVP in all four scenes and observed a significant decrease of performance in the "intersection" scene, with an increase in mean per MPJPE to 22.62cm and a recall rate of just 77.39\%. We attribute this primarily to differences in human pose distribution between scenarios, particularly the limited availability of human pose in intersection scenes due to the prevalence of vehicles in street view imagery and the resulting small proportion of detected pedestrians. As a consequence, the algorithm was not well-trained to handle this type of data, which is a minor component of the overall dataset. This highlights the challenge of database distribution disparities across different scenes of the HumanCT-3D dataset.

\begin{figure*}
  \centering
  \begin{subfigure}{0.22\linewidth}
  \includegraphics[width=1.0\textwidth]{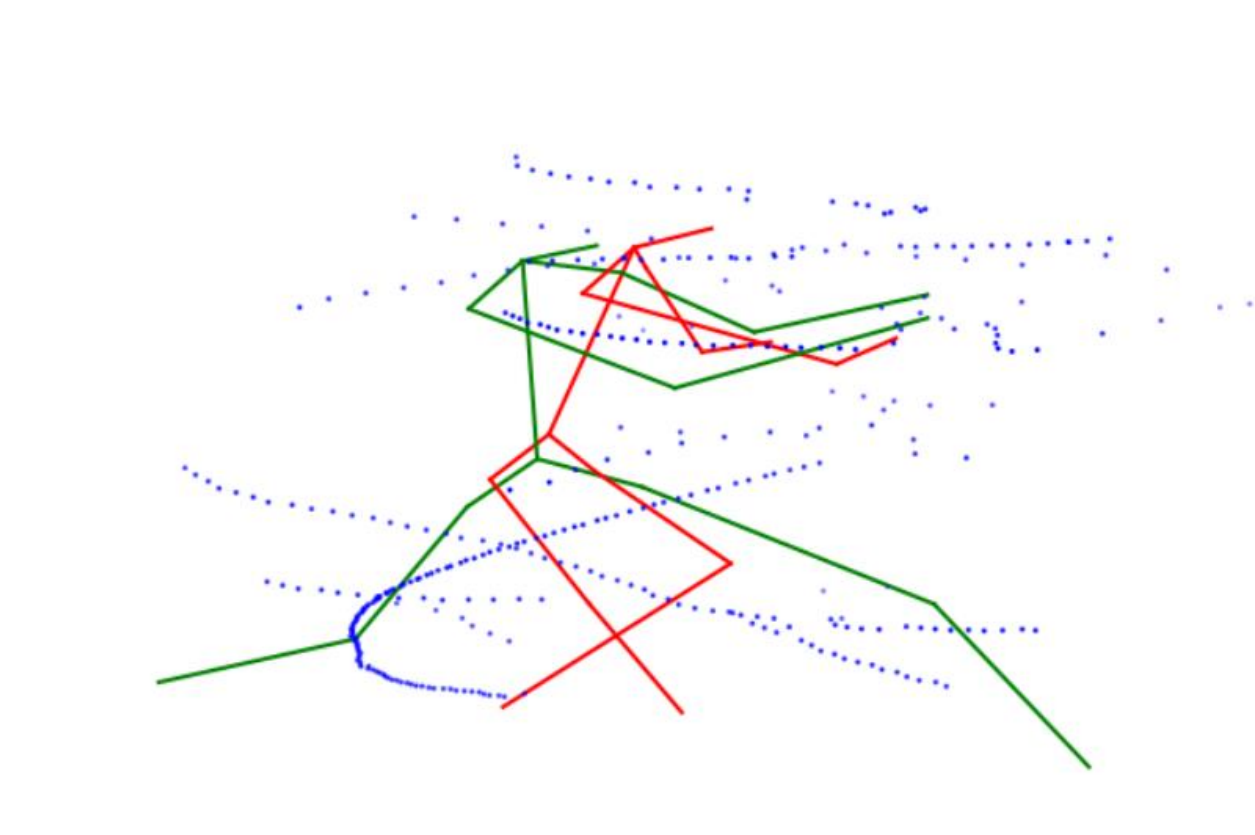}
  \subcaption*{}
    \label{fig:failpcd1}
  \end{subfigure}
  \begin{subfigure}{0.18\linewidth}
  \includegraphics[width=1.0\textwidth]{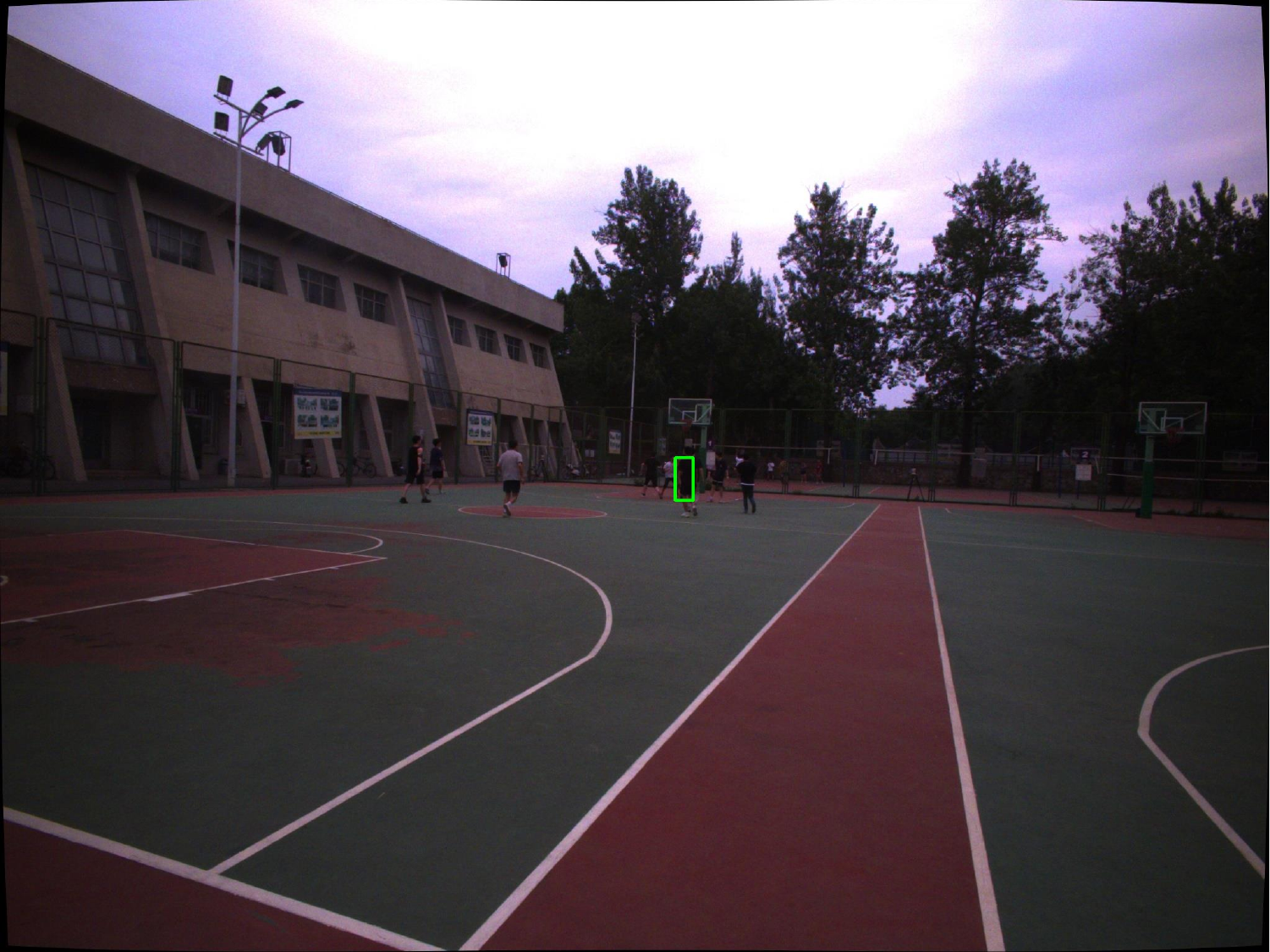}
  \subcaption*{}
    \label{fig:failcam01}
  \end{subfigure}
  \begin{subfigure}{0.18\linewidth}
  \includegraphics[width=1.0\textwidth]{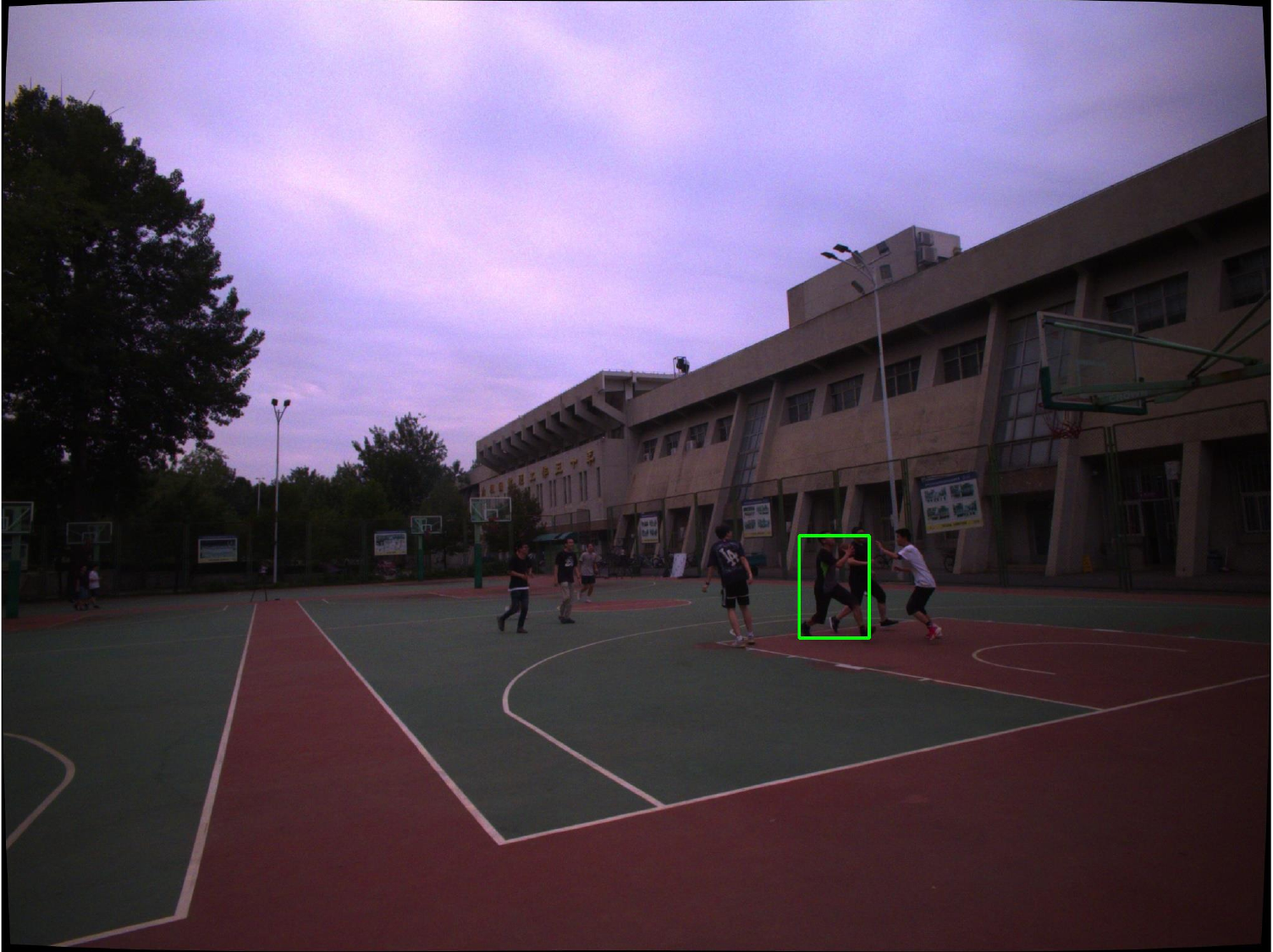}
  \subcaption*{}
    \label{fig:failcam11} 
  \end{subfigure}
 \begin{subfigure}{0.18\linewidth}
  \includegraphics[width=1.0\textwidth]{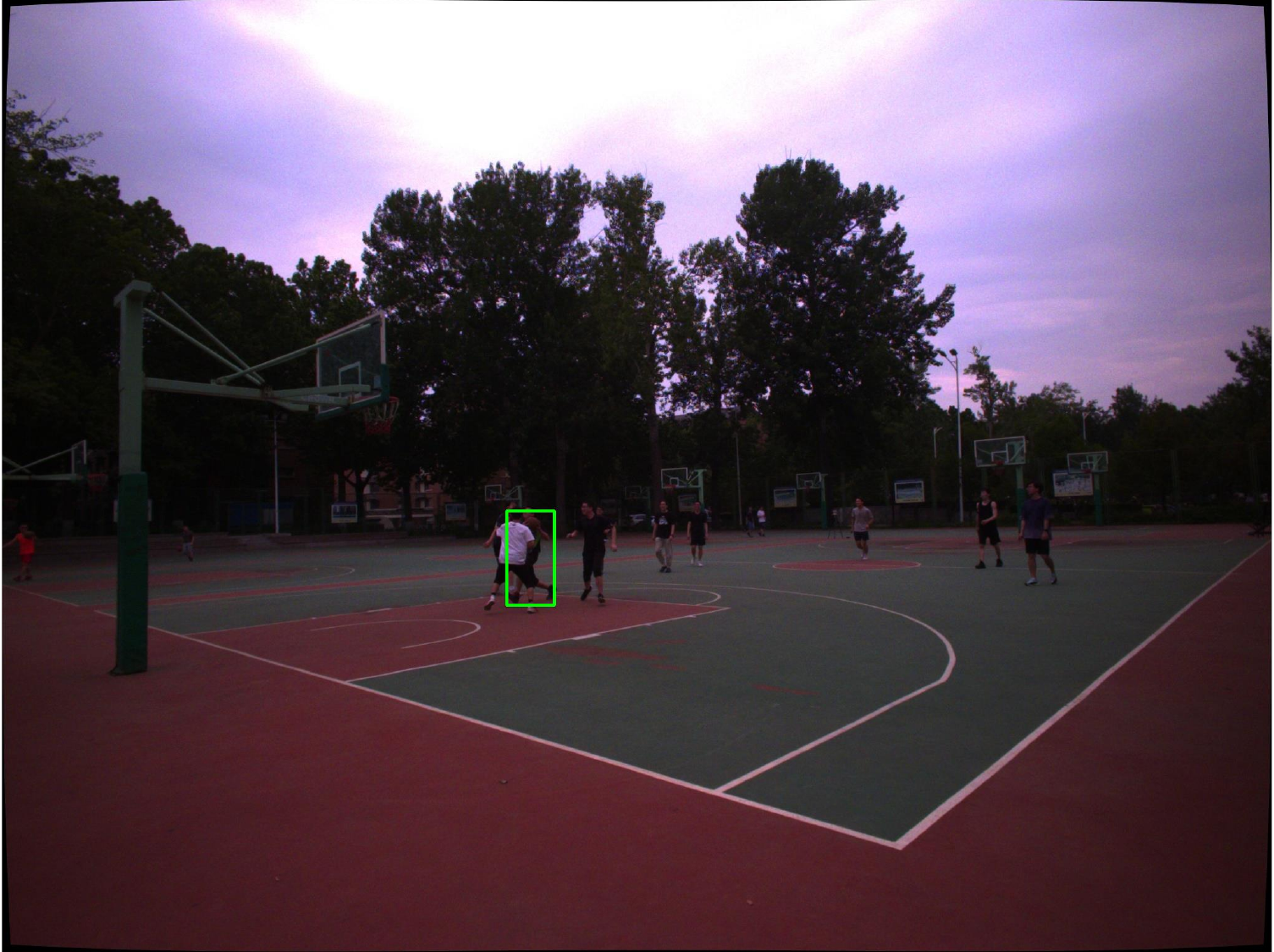}
  \subcaption*{}
    \label{fig:failcam21}
  \end{subfigure}
  \begin{subfigure}{0.18\linewidth}
  \includegraphics[width=1.0\textwidth]{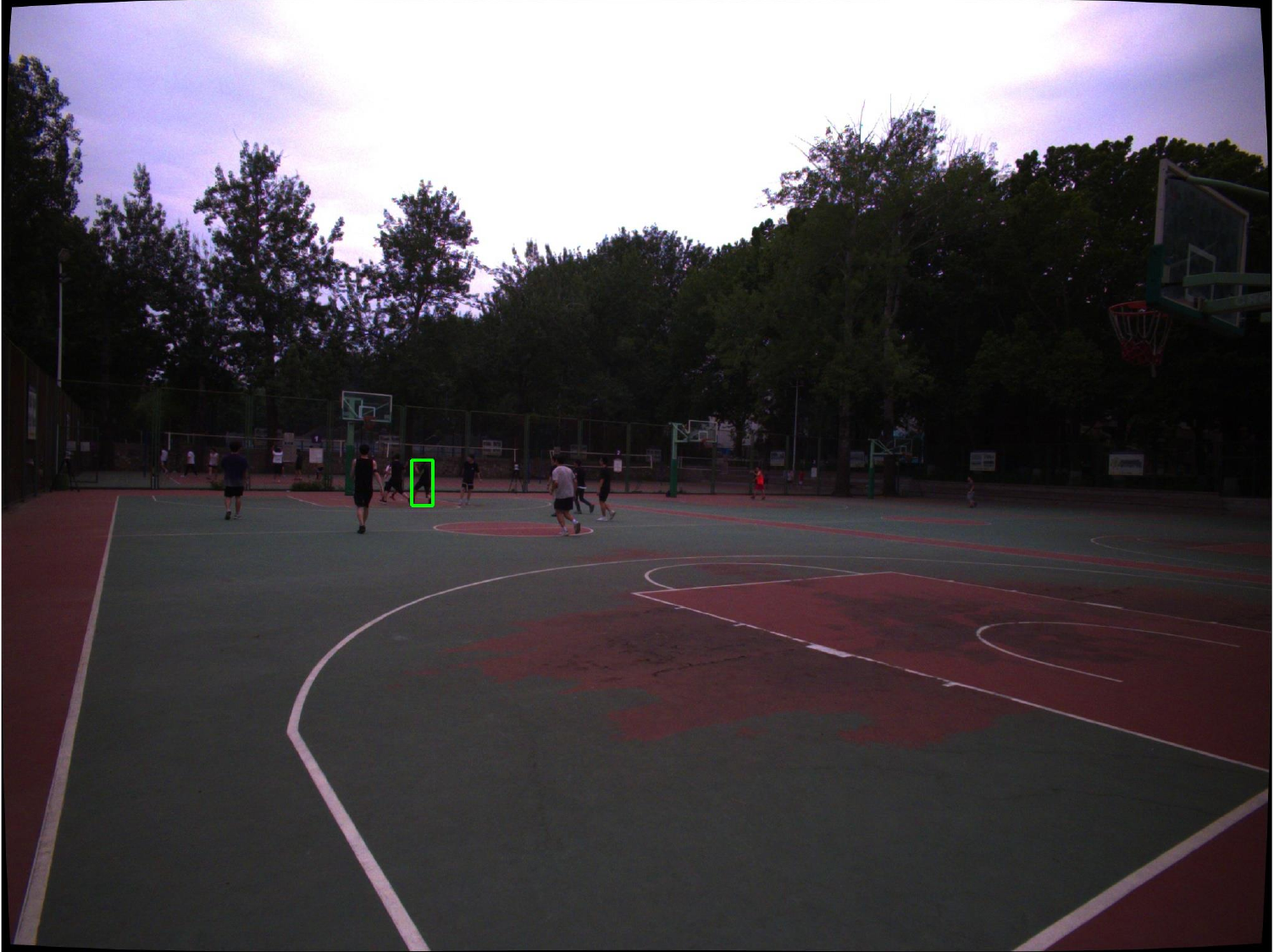}
  \subcaption*{}
    \label{fig:failcam31} 
  \end{subfigure}

  \begin{subfigure}{0.22\linewidth}
  \includegraphics[width=1.0\textwidth]{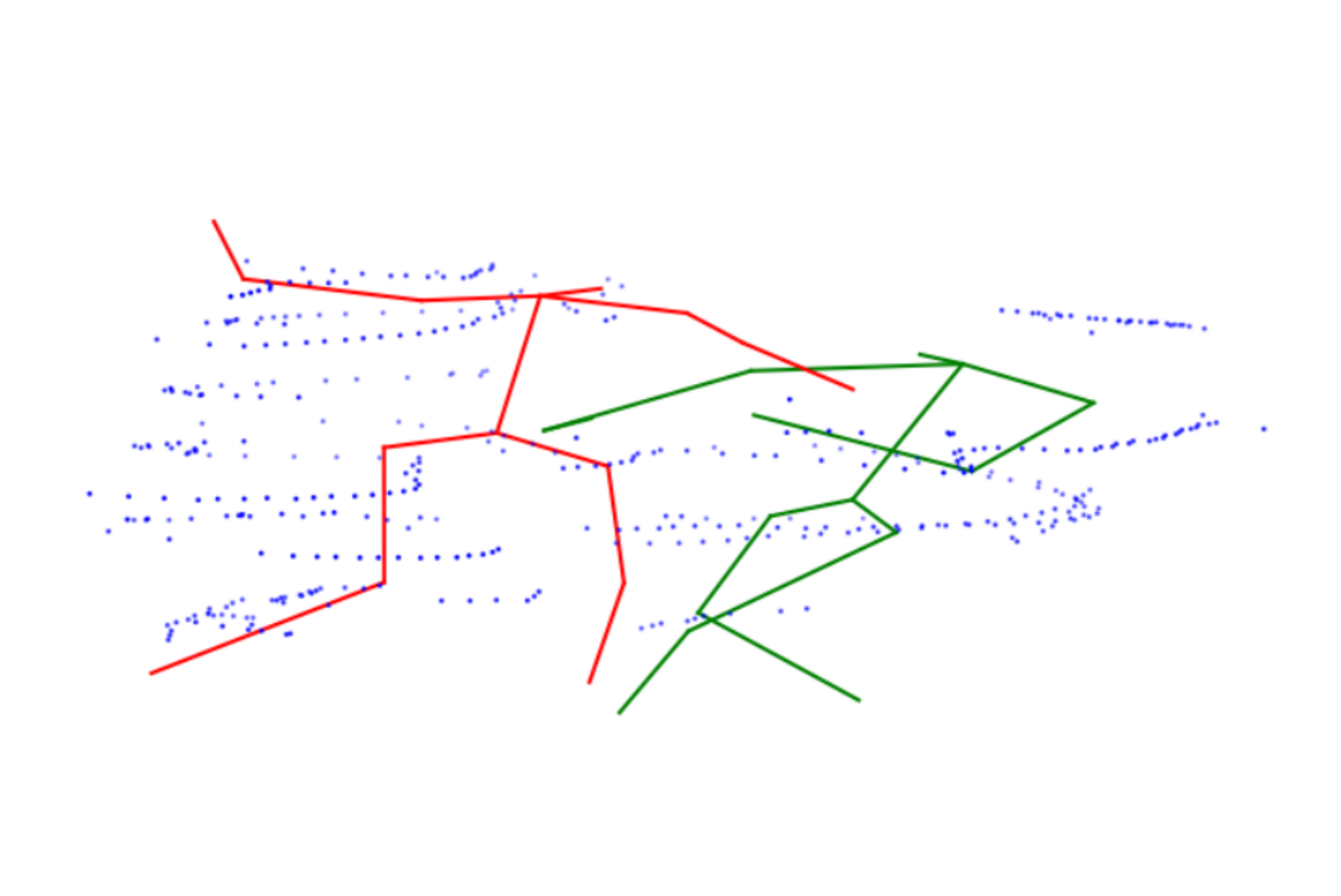}
  \subcaption*{}
    \label{fig:failpcd2}
  \end{subfigure}
  \begin{subfigure}{0.18\linewidth}
  \includegraphics[width=1.0\textwidth]{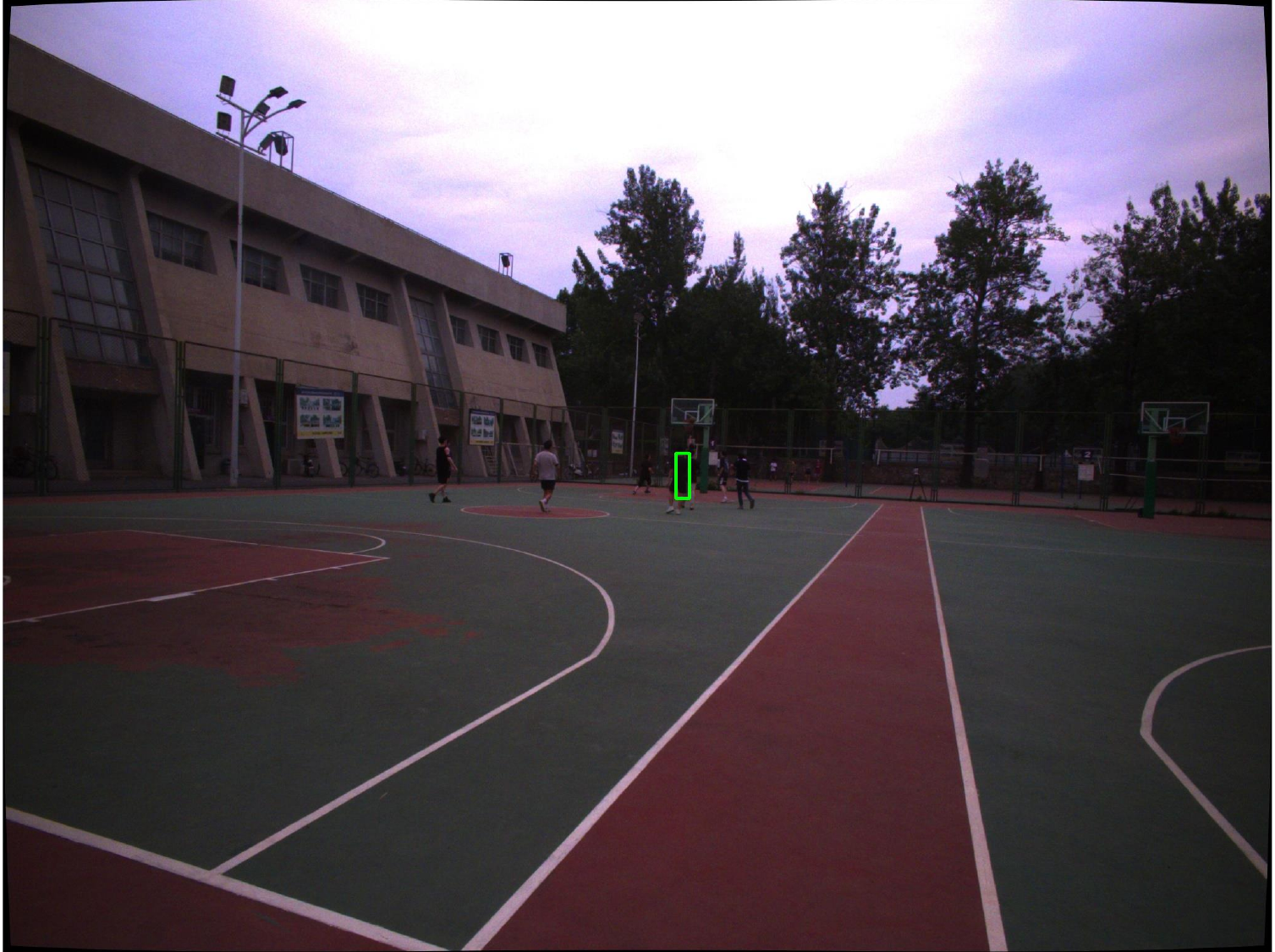}
  \subcaption*{}
    \label{fig:failcam02}
  \end{subfigure}
  \begin{subfigure}{0.18\linewidth}
  \includegraphics[width=1.0\textwidth]{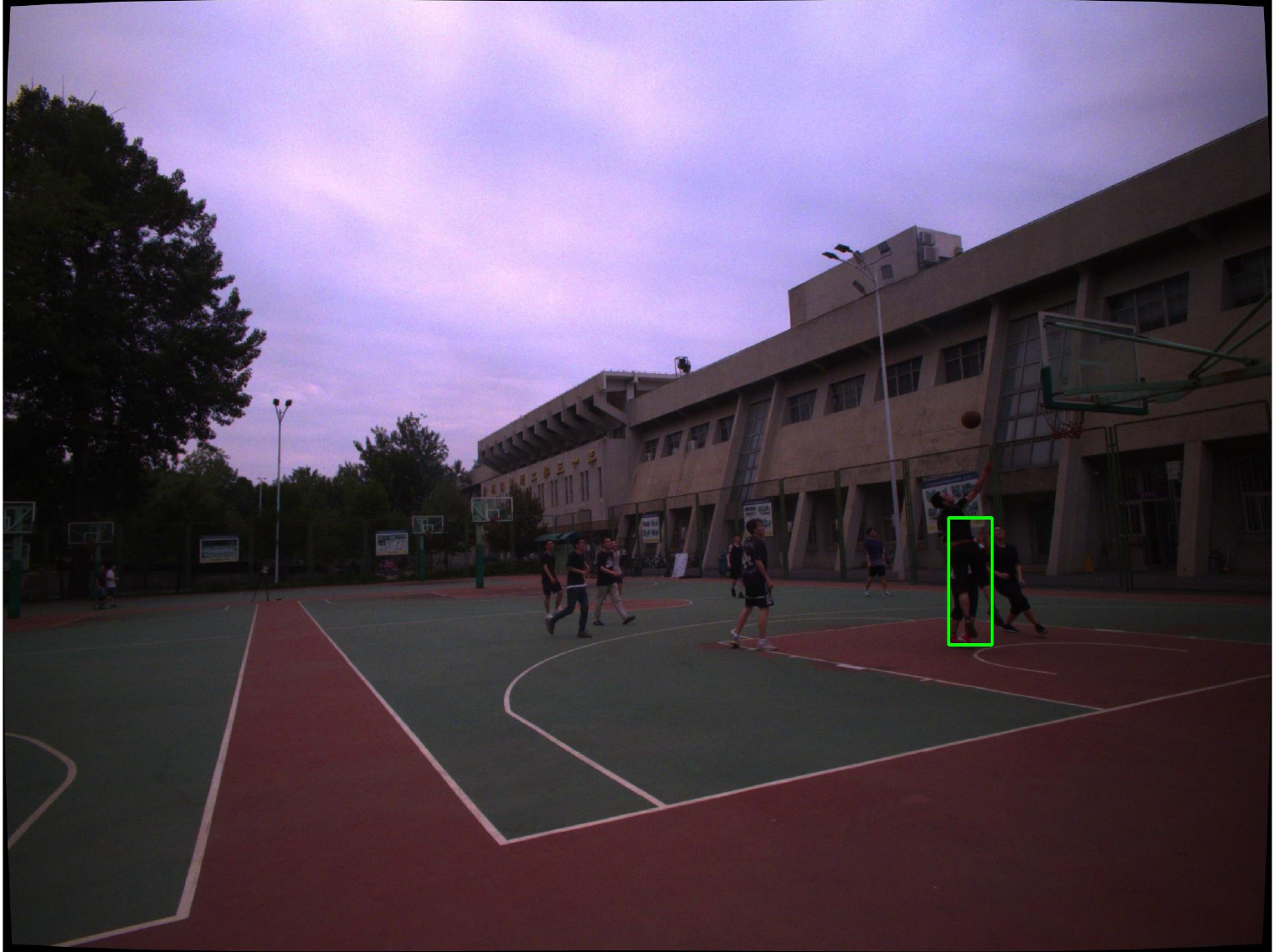}
  \subcaption*{}
    \label{fig:failcam12} 
  \end{subfigure}
 \begin{subfigure}{0.18\linewidth}
  \includegraphics[width=1.0\textwidth]{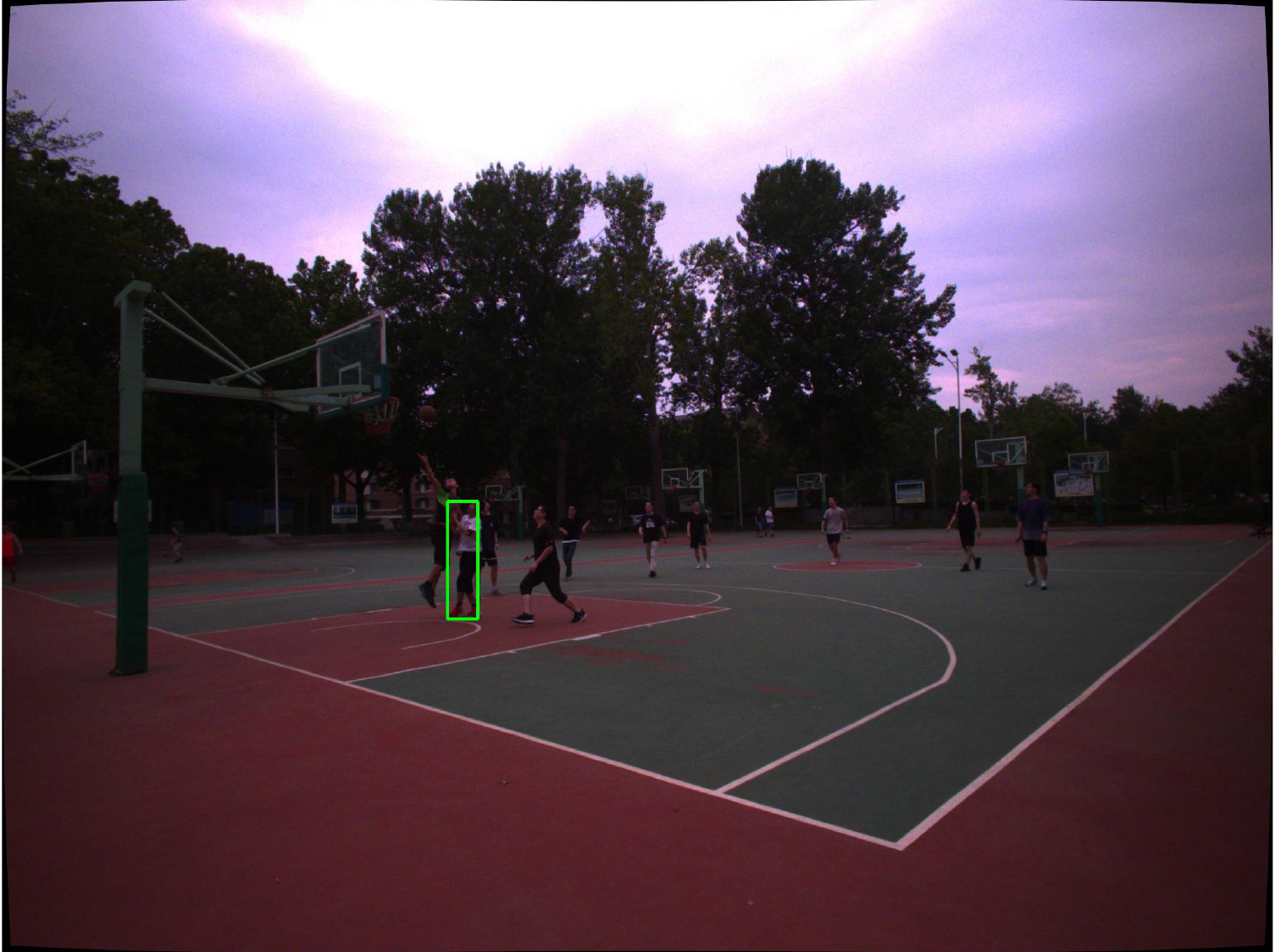}
  \subcaption*{}
    \label{fig:failcam22}
  \end{subfigure}
  \begin{subfigure}{0.18\linewidth}
  \includegraphics[width=1.0\textwidth]{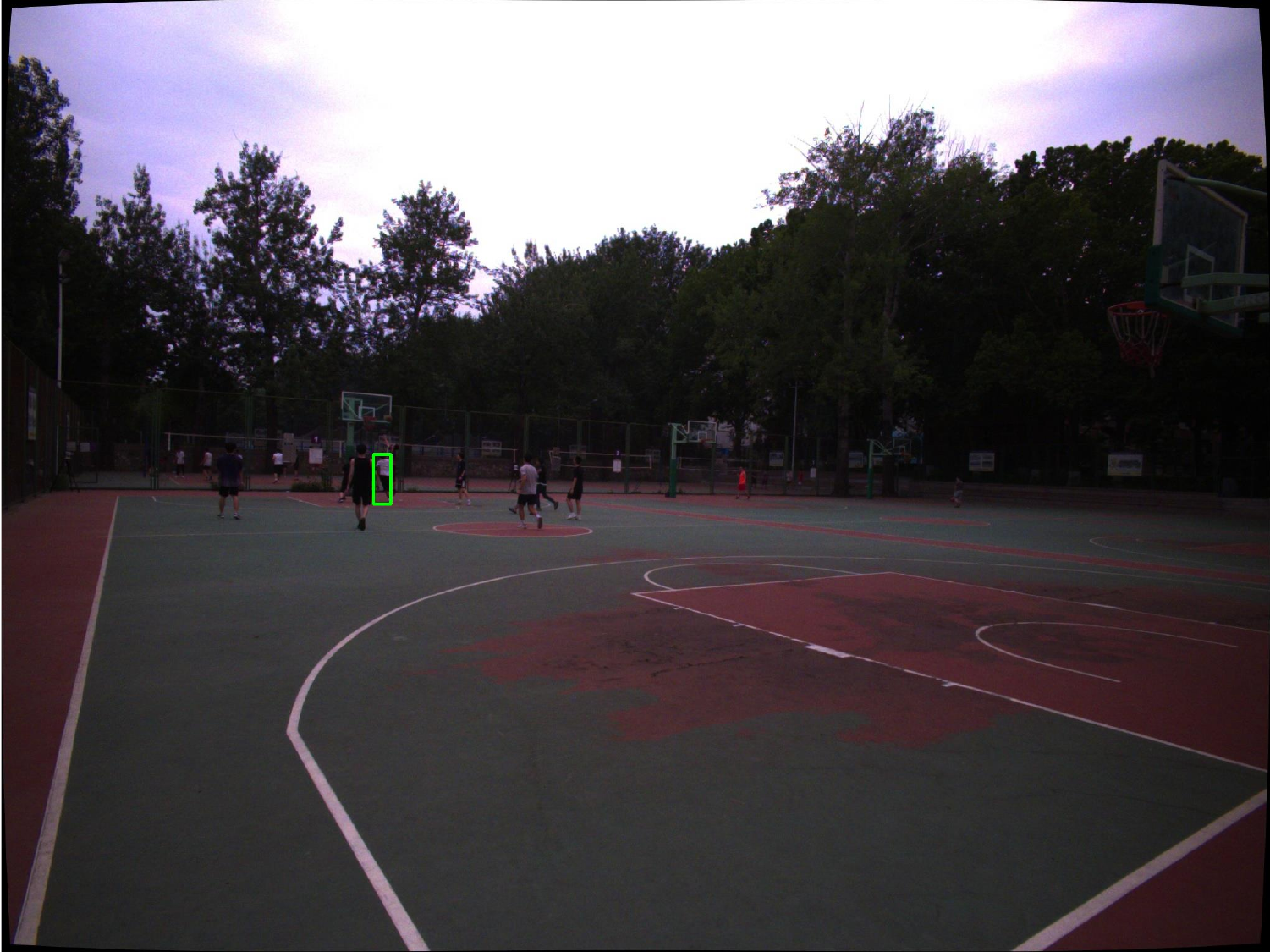}
  \subcaption*{}
    \label{fig:failcam32} 
  \end{subfigure}
  \vspace{-7mm}
  \caption{Two failure cases. The first row shows a failure case caused by heavy occlusion and long distance between human and camera. The second row exhibits a failure case attributed to the proximity of two individuals. First column shows predicted 3D human pose (red) and ground truth (green) overlapped with local pointclouds. Other columns show the corresponding RGB images, with green bounding boxes mark the corresponding 2D position of the human in images.}
  \label{fig:failure}
  \vspace{-5mm}
\end{figure*}

\textbf{Failure cases and analysis.}
Fig. \ref{fig:failure} shows two failure cases, which also represent the most frequent case. One is limited information, as shown in the first row of Fig. \ref{fig:failure}. This person is too far away from the two sensors in the scene and is heavily obscured in the third view, so only one sensor is useful. In addition, pointcloud is sparse due to occlusion, which leads to the failure of the algorithm. The other is proximity of two individuals, this not only causes them to occlude each other in camera view, but also causes their point clouds to coincide with each other, which is quite challenging . The existence of similar samples increases the difficulty and richness of the Human-M3 dataset.

\section{Limitations}
Although multiple-view scenes are commonly regarded as the optimal setting for capturing human motion, the configuration and calibration processes involved are both unwieldy and time-intensive. Furthermore, although the proposed optimization algorithm prioritized efficiency in both spatial and temporal domains, it did not integrate the VAE methodology detailed in \cite{rempe2021humor}. To further improve labeling accuracy and leverage stronger motion priors, implementation of a temporal learning approach may prove beneficial, but manual intervention and screening remain necessary at this stage. It is also important to acknowledge that the limited range of scenes and human motion patterns covered in this study may not fully represent the diversity of real-world scenarios, thereby underscoring the need for more comprehensive data collection strategies. Lastly, while initial findings demonstrate promising results, further investigation into more advanced fusion techniques is necessary to unlock the full potential of multi-modal datasets like Human-M3.

\section{Conclusion}
This article introduces an outdoor human pose database named Human-M3, which possesses multiple views and modalities of multiple persons in varying scenarios. The variety and richness of the scenes and human poses are noteworthy. We developed an optimization method based on the multi-modal data, which is efficient and yields accurate results for 3D human pose annotation. Moreover, we propose a multi-modal based algorithm MMVP for combining different modalities of input, which serves as a baseline for multi-modal 3D human pose estimation algorithms and highlights the value of multi-modal data in 3D human pose estimation.

\begin{appendices}
\section{Detailed descriptions of some parts of the main text}
\subsection{Camera extrinsic parameter calibration.}
In the traditional multi-view video collection process, determination of the extrinsic matrix of the camera can frequently be proved challenging due to the difficulty that arises in establishing correspondence between the RGB image keypoints and their corresponding real-world 3D coordinates. Multimodal data solves this issue. We simply used the LiDAR coordinate system as the world coordinate system and manually annotated coordinate pairs $P = (P_{W}, P_{I})$, where the coordinates of the actual location were consistent in both pointcloud and 2D image space. Based on camera projection principle, 3D points could be projected into 2D image space following:
\begin{equation}
\label{proj_formula}
 [u\ v\ 1]^T=IE[X\ Y\ Z\ 1]^T 
\end{equation}
\begin{equation}
 P_{p} = [u,v]
\end{equation}

where $[u\ v ]$ is the 2D image pixel coordinate, $[X\ Y\ Z]$ is the 3D world coordinate, $I$ and $E$ are the intrinsic and extrinsic matrices, respectively. 

\begin{figure*}
  \centering
  \begin{subfigure}{0.22\linewidth}
  \includegraphics[width=1.0\textwidth]{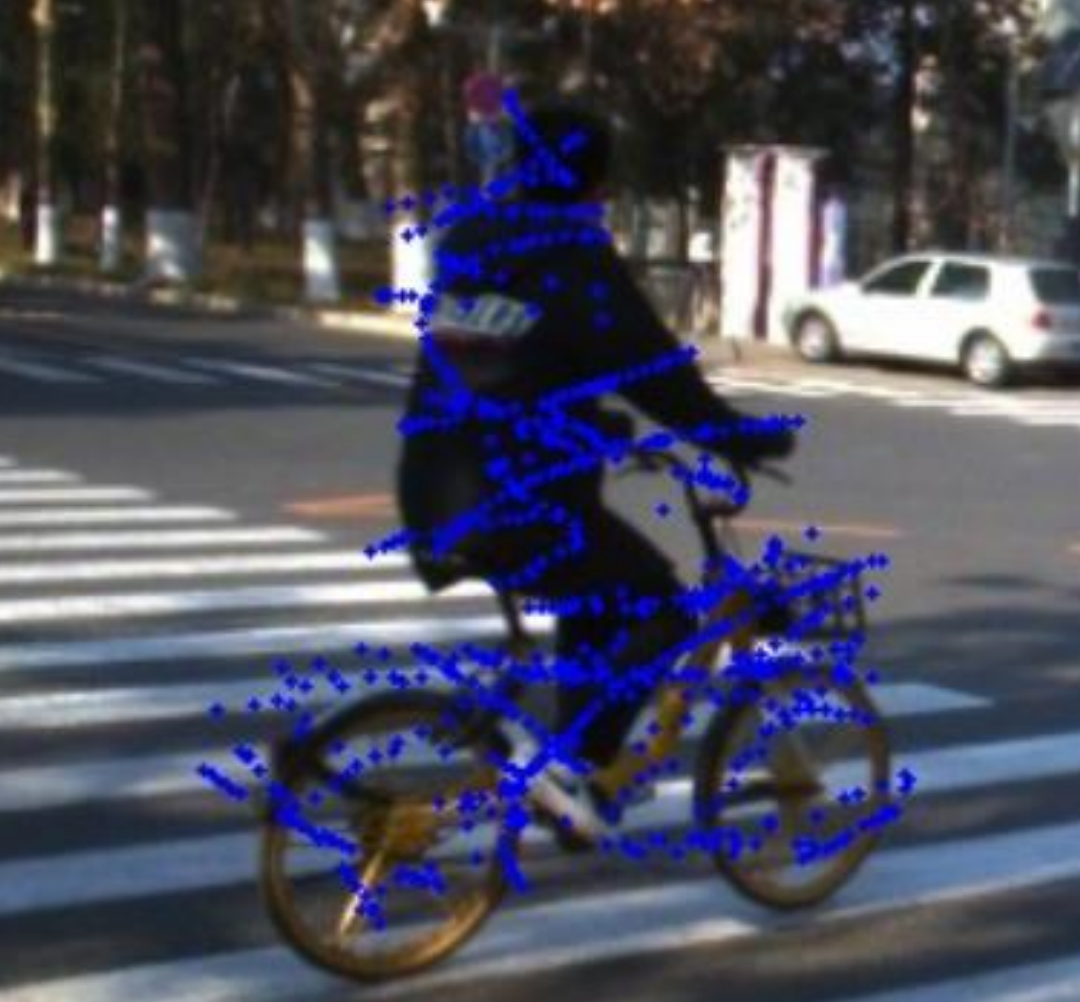}
  \subcaption*{}
  \end{subfigure}
  \begin{subfigure}{0.22\linewidth}
  \includegraphics[width=1.0\textwidth]{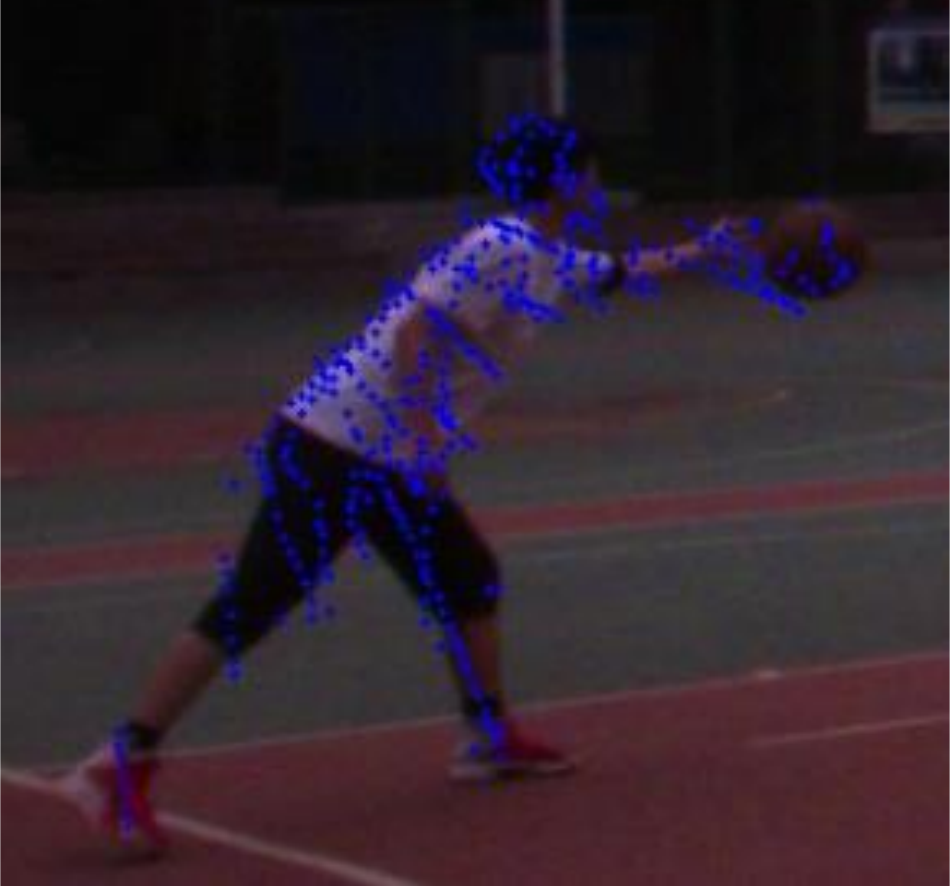}
  \subcaption*{}
  \end{subfigure}
  \begin{subfigure}{0.22\linewidth}
  \includegraphics[width=1.0\textwidth]{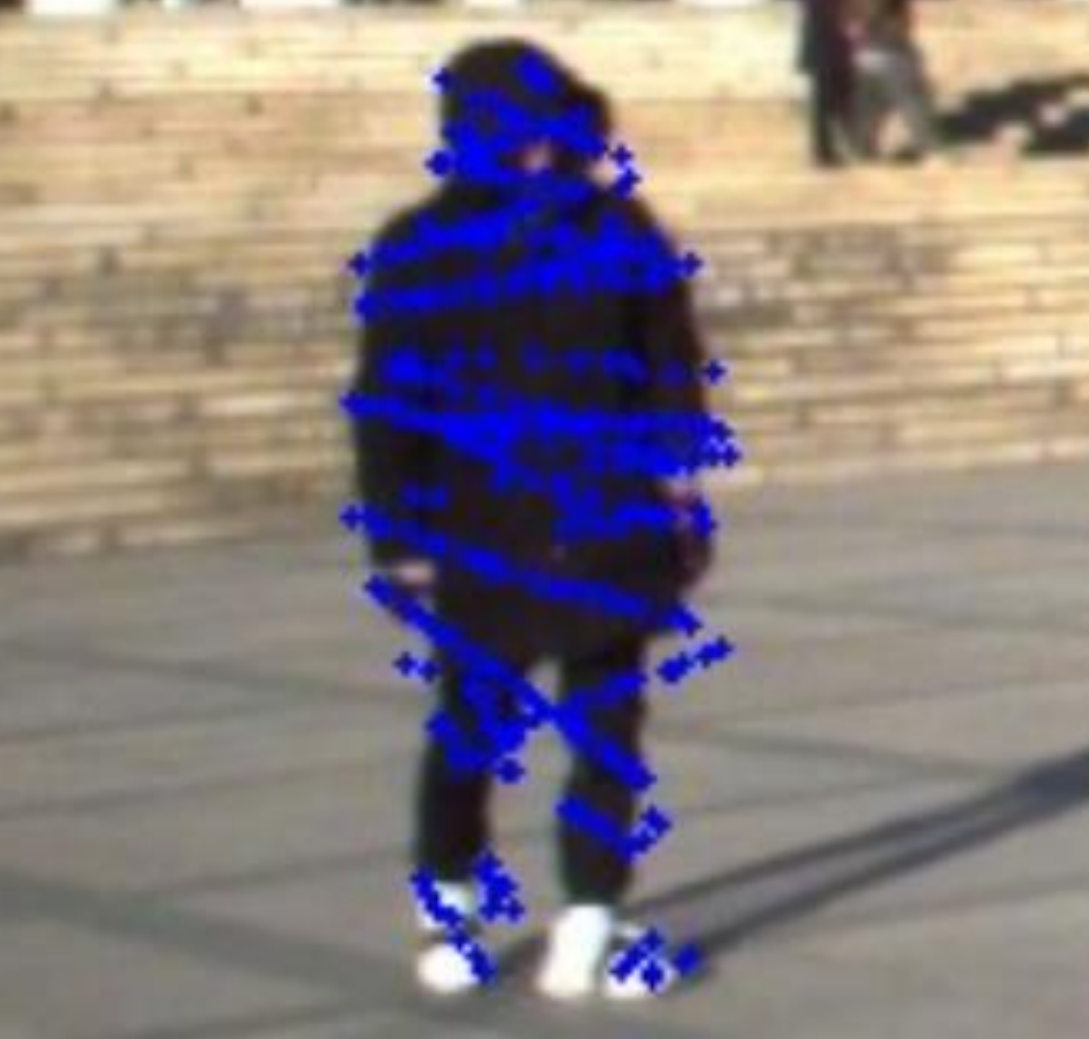}
  \subcaption*{}
  \end{subfigure}
  \begin{subfigure}{0.22\linewidth}
  \includegraphics[width=1.0\textwidth]{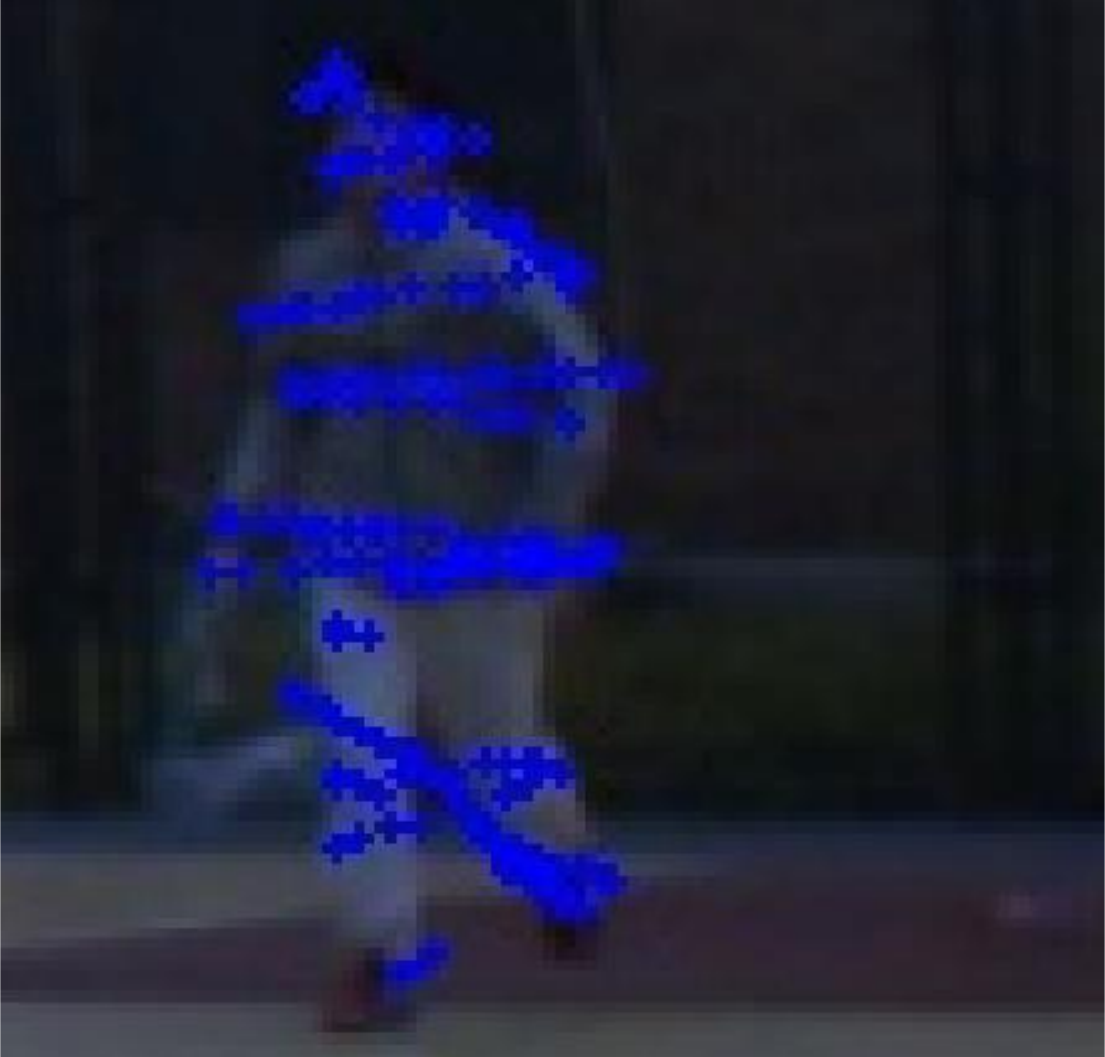}
  \subcaption*{}
  \end{subfigure}
  \vspace{-4mm}
  \caption{Examples from four scenes after camera parameter calibration.}
  \vspace{-6mm}
  \label{fig:camera_parameter}
\end{figure*}

We then used gradient descent to fit the extrinsic matrices for minimizing the loss function, which is defined as $l_2$-norm between the projected 2D coordinates of annotated 3D points and 
corresponding annotated 2D coordinates: 
\begin{equation}
L=\frac{1}{N}\sum_{k=1}^{N}||P_{p}^{k}-P_{I}^{k}||_2^2
\end{equation}
where $N$ is the total number of annotated coordinate pairs in pointcloud and image. It is worth noting that in order to ensure the accuracy of the camera parameters over the entire spatial range, we tried to select annotated points in a scattered manner, and ensured that any three annotated points are not on the same plane in 3D space. As illustrated in Fig. \ref{fig:camera_parameter}, the camera parameters fit well with the pointcloud, and the pointcloud can also fit the edge contour of the human body well. It is worth mentioning that since we only demarcated 3D points in the field, camera parameters may still have several pixel errors at the edge of the field, but few pedestrian samples were affected by this problem.

\subsection{Details about temporal human pose optimization. }
\textbf{Priors.} SMPL \cite{loper2015smpl} defined a set of deformation priors describing human meshes, The deformation law of human surface is described by a well-trained model which is driven by 10-dim shape parameters $\beta$ and 72-dim pose parameters. So in the text we estimate the shape and pose of the human body by estimating this set of parameters as well as humans' global location. In the follow-up work of SMPL, smplify-x \cite{pavlakos2019expressive} used a motion-capture database AMASS \cite{mahmood2019amass} which includes amounts of captured human poses to learn valid human pose and shape prior based on VAE, called VPoser. Specifically, VPoser established a well-learned MLP to transform the 72-dim pose parameter into 32 dimensions, or vice versa.
These two parts of the prior well constrained the rationality of human poses.

\textbf{Loss definition.}
Equation (1), (2) and (3) in the main text gives a brief description about effects of various parts of the loss, but does not detail their form. Specifically, $L_{sp}(\beta)$ is defined as the two-norm of 10-dim shape parameter $\beta$, and $L_{pp}(\theta)$ is defined as the two-norm of 32-dim latent vector following VPoser. Other parts of the prior loss is defined as:
\begin{equation}
 P_{f}^{3d}, M_f = S\_(r_f, \beta, {\theta}_f) (f = 0,...,t-1)
\end{equation}
\begin{equation}
 L_I(r,\beta,\theta,P) = \sum_{f=0}^{t-1} \sum_{c=1}^{C} ||IE_c P_{f}^{3d} - P_{cf}||^2_2
\end{equation}
\begin{equation}
 L_P(r,\beta,\theta,Q) = \sum_{f=0}^{t-1} D_{ch}(M_f, Q_f)
\end{equation}
where $S\_$ indicates SMPL model inference, $M_f$ indicates the 3D mesh points describing the whole human mesh and $P_{f}^{3d}$ indicates the 3D human pose joints processed by SMPL model. $IE_c$ indicates the projection process described in equation \ref{proj_formula} for camera $c$. $D_{ch}$ indicates the chamfer distance loss. The definition of $Q_f$ and $P_{cf}$ is consistent with that in equation (1), (2) and (3) in the main text.

For motion smooth loss $L_{mp}(\theta)$, we have:
\begin{equation}
 L_{mp}(\theta) = L_{o}(\theta) + L_{p}(\theta) + L_{j}(\theta)
\end{equation}
\begin{equation}
 L_{p}(\theta) = \sum_{f=1}^{t} F_e(|| \theta_f - \theta_{f-1} ||^2_2, T_p)
\end{equation}
\begin{equation}
 L_{j}(\theta) = \sum_{f=1}^{t} F_e(|| P^{3d}_f - P^{3d}_{f-1} ||^2_2, T_j)
\end{equation}
\begin{equation}
 L_{o}(\theta) = \sum_{f=1}^{t} F_e(|| Ro_f - Ro_{f-1} ||^2_2, T_o).
\end{equation}
where $F_e$ is the exponential loss preventing too large excessive variables:
\begin{equation}
 F_e(x,y) = \left \{
 \begin{array}{ll}
    e^{\frac{x}{y}-1} - 1,                    & y \leq x \leq 20y\\
    0,                                 & otherwise,
\end{array}
\right.
\end{equation}
and $Ro_f$ is the corresponding root orient vector derived from $M_f$ and $T_p, T_j, T_o$ are hyper-parameters which vary with scenes.

\begin{figure*}
  \centering
  \includegraphics[width=0.8\textwidth]{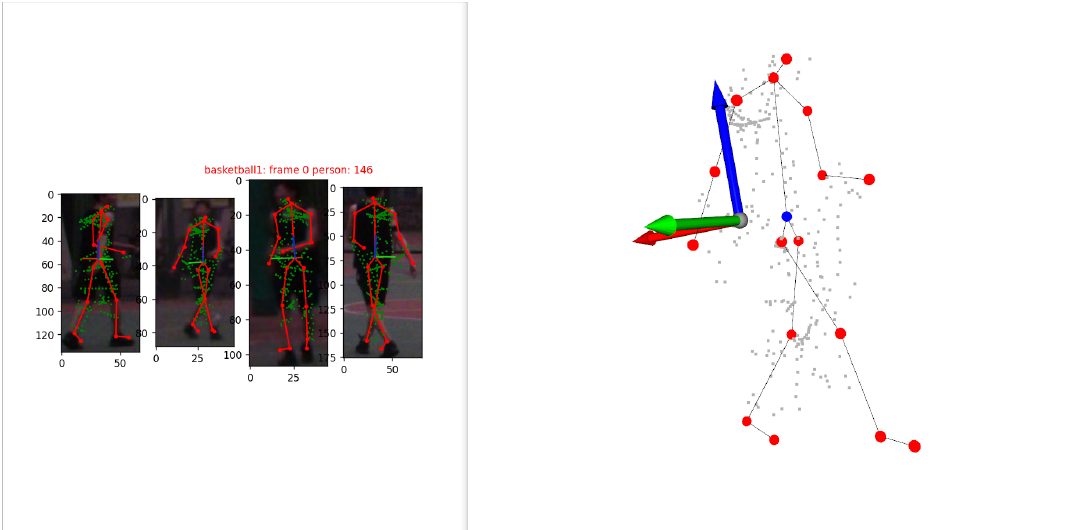}
  \vspace{-2mm}
  \caption{Calibration tool for test set human pose annotation. Left: Images captured from different viewpoints of the same person, along with their corresponding poses. The green dots represent the projection of the 3D point cloud. Right: The corresponding pedestrian point cloud and three reference directions of the world coordinate. The blue dot indicates the keypoint under annotation.}
  \label{fig:calib_pose}
  \vspace{-3mm}
\end{figure*}

\begin{figure*}
  \includegraphics[width=1.0\textwidth]{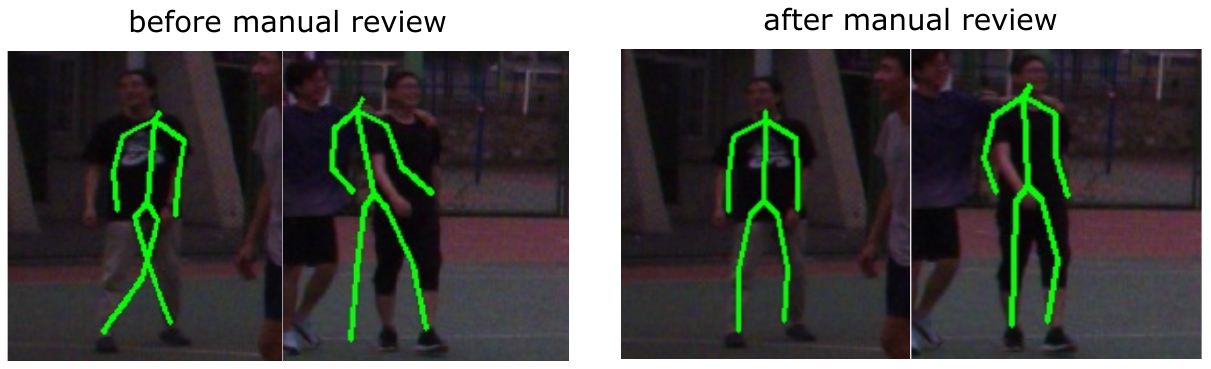}
  \caption{Two examples of human pose annotation before and after manual review. The two pictures on the left show random pose error occurred during optimization. The two pictures on the right show revised poses after manual review, which is obtained by automatic interpolation step.}
  \label{fig:manual}
\end{figure*}

\begin{figure*}
  \includegraphics[width=1.0\textwidth]{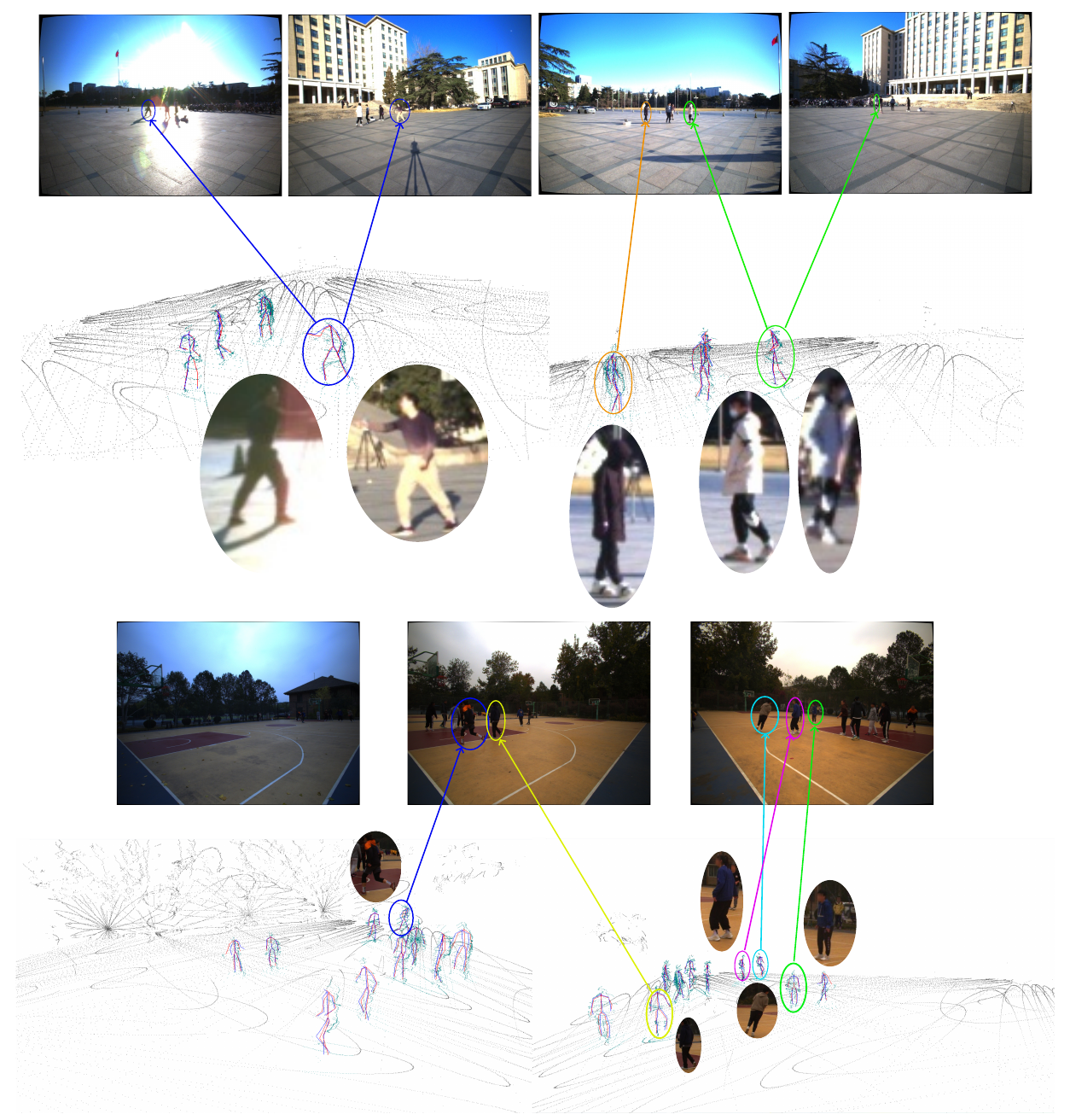}
  \caption{Two exmples of MMVP results on the proposed Human-M3 dataset. In the second row, side views of the two scenes are shown separately. Inside them, red lines indicate predicted 3D human poses and blue lines indicate ground truth poses. Foreground points are marked as cyan points, and background points are gray. Corresponding RGB images are shown in the first row. We use ellipses and arrows to mark the correspondence for better understanding.}
  \label{fig:algo_show_more}
\end{figure*}
Finally, in order to ensure the reliability of the test set, we manually annotated all the poses of the test set (the last 10\% frames of each scene) using multi-view epipolar geometry constraints and the consistency information from multi-modal data. As shown in Fig. \ref{fig:calib_pose}, with simultaneous knowledge of the projection relationships from various viewpoints and the corresponding 3D point cloud, annotators can determine the 3D pose of humans more accurately and efficiently.

\section{More figures}
Fig. \ref{fig:manual} illustrates some examples for comparison to highlight the importance of manual review mentioned in section 3.2 in the main text. It took about 25 hours for a single person in manual review process for the whole Human-M3 dataset. And Fig. \ref{fig:algo_show_more} illustrates two more examples of proposed MMVP baseline for 3D HPE in Human-M3 dataset in different scenes.
\end{appendices}

\bibliographystyle{plain}

\end{document}